\documentclass[journal]{IEEEtran}
\usepackage{grffile}
\usepackage[utf8]{inputenc} 
\usepackage{hyperref}
\newcommand{\coloredcite}[1]{{\color{blue}\cite{#1}}}
\hypersetup
{
	colorlinks=true,
	linkcolor=blue,
	filecolor=blue,
	urlcolor=blue,
	citecolor=blue,
}
\usepackage{amsmath,amsfonts}
\usepackage{algorithmic}
\usepackage{algorithm}
\usepackage{array}
\usepackage{textcomp}
\usepackage{stfloats}
\usepackage{url}
\renewcommand{\arraystretch}{1.5}
\usepackage{verbatim}
\usepackage{graphicx}
\graphicspath{{./}}
\usepackage{cite}
\usepackage{bm}
\usepackage{cleveref}
\usepackage{amssymb}
\usepackage{threeparttable}
\usepackage{float}
\usepackage{booktabs}
\usepackage{multirow}
\usepackage{diagbox}
\usepackage{subfigure}
\usepackage{bbm}
\usepackage{makecell}

\ifCLASSINFOpdf

\else

\fi

\begin{document}

\title{Few-shot Unknown Class Discovery of Hyperspectral Images with Prototype Learning and Clustering}

\author{Chun Liu, Chen Zhang, Zhuo Li, Zheng Li, Wei Yang

\thanks{Chun Liu, Chen Zhang, Zhuo Li, Zheng Li and Wei Yang are with School of Computer and Information Engineering, Henan Key Laboratory of Big Data Analysis and Processing, Henan Engineering Laboratory of Spatial Information Processing and Henan Industrial Technology Academy of Spatio-Temporal Big Data, Henan University, Zhengzhou 450046, China (e-mail: liuchun@henu.edu.cn; ac18236911774@henu.edu.cn; a13273973603@henu.edu.cn; lizheng@henu.edu.cn; weiyang@henu.edu.cn).}

\thanks{Manuscript received April 19, 2005; revised August 26, 2015.}}

\markboth{}%
{Shell \MakeLowercase{\textit{et al.}}: Bare Demo of IEEEtran.cls for IEEE Journals}

\maketitle

\begin{abstract}
   Open-set few-shot hyperspectral image (HSI) classification aims to classify image pixels by using few labeled pixels per class, where the pixels to be classified may be not all from the classes that have been seen. To address the open-set HSI classification challenge, current methods focus mainly on distinguishing the unknown class samples from the known class samples and rejecting them to increase the accuracy of identifying known class samples. They fails to further identify or discovery the unknow classes among the samples. This paper proposes a prototype learning and clustering method for discoverying unknown classes in HSIs under the few-shot environment. Using few labeled samples, it strives to develop the ability of infering the prototypes of unknown classes while distinguishing unknown classes from known classes. Once the unknown class samples are rejected by the learned known class classifier, the proposed method can further cluster the unknown class samples into different classes according to their distance to the inferred unknown class prototypes. Compared to existing state-of-the-art methods, extensive experiments on four benchmark HSI datasets demonstrate that our proposed method exhibits competitive performance in open-set few-shot HSI classification tasks. All the codes are available at  \href{https://github.com/KOBEN-ff/OpenFUCD-main} {https://github.com/KOBEN-ff/OpenFUCD-main}

\end{abstract}


\IEEEpeerreviewmaketitle

\section{Introduction}

\IEEEPARstart{H}{yperspectral} imaging (HSI) captures the spatial and spectral information of objects simultaneously. The detected spectral information far exceeds human visual perception, enhancing our understanding of the natural world \coloredcite{r1}\coloredcite{r2}. This leads to the widespread and successful application of HSIs in the fields of environmental science, agriculture, ecology, marine science, geology, and land management\coloredcite{r3}. 

HSI classification is to infer the classes of the pixels in the images and accordingly recognize the different objects captured \coloredcite{r29},\coloredcite{r30},\coloredcite{r31},\coloredcite{r32}. Early methods for HSI classification predominantly focused on the rich spectral features embedded within each pixel, and  utilized conventional classifiers such as Random Forest (RF)\coloredcite{r58}, K-Nearest Neighbors (KNN)\coloredcite{r59}, Support Vector Machines (SVM)\coloredcite{r60}, and Multinomial Logistic Regression\coloredcite{r61}  directly on the spectral features. They sometimes also employed dimensionality reduction techniques like Principal Component Analysis (PCA)\coloredcite{r62} and Linear Discriminant Analysis (LDA)\coloredcite{r63} to alleviate the redundancy in high-dimensional spectral data. These methods overlooked spatial information, relying solely on the spectral information of pixels, which limited their classification performance. 

In recent years, deep learning  models such as convolutional neural networks (CNNs) and recurrent neural networks (RNNs) have been employed to capture spatial-spectral features, enhancing both the accuracy and robustness of HSI classification. These models have facilitated the development of a range of innovative methods for HSI classification. Particularly, due to the practical limitations of human monitoring over large geographic areas and the high costs of expert annotations, HSI classification based on few-shot learning \coloredcite{r54},\coloredcite{r55},\coloredcite{r56},\coloredcite{r57} is favored for its recognition capabilities under the setting that there are few labeled samples available\coloredcite{r64}, \coloredcite{r65}. It often obtains prior knowledge from the classes where there are a large number of labeled samples, and effectively transfer them to target classes where there are only few labeled samples\coloredcite{r5}.

However, most of current deep learning methods for HSI classification, including the few-shot learning methods, are based on the closed-set assumption that all the samples to be classified come from the same classes that have been seen in the training stage. This assumption does not holds in reality. For example, due to the diversity of land cover (small land covers may be easily overlooked during manual labeling) and the emergence of new land cover classes over time, the classifier will encounter samples with classes that are unknown before. In closed-set hyperspectral classification, linear classification layers or the softmax function are commonly used, which will misclassify unknown class samples into one of the known classes, significantly increasing the risk of misclassification in open-set environments. Therefore, compared to traditional closed-set HSI classification, open-set HSI classification not only needs to accurately classify known class samples but also must have the capability to identify unknown class samples. 

To address HSI classification challenges in open-set or  open-world environments, many methods have been presented. Some discriminative methods  applied confidence thresholds which were typically determined based on factors such as the distribution of decision scores \coloredcite{r26} and Extreme Value Theory (EVT)\coloredcite{r24}, \coloredcite{r22}, to reject samples with low classification confidence and thereby identify unknown class instances. For example, Liu et al.\coloredcite{r15} and Yue et al.\coloredcite{r22} proposed the method to segregate unknown classes from known classes and classify known class samples through the reconstruction of HSI spectral and spatial features. Spectral-spatial reconstruction error was modeled using Extreme Value Theory (EVT) and Weibull distribution to delineate boundaries between known and unknown classes. Larger reconstruction errors were considered indicative of unknown classes. Pai et al. introduced an outlier calibration network\coloredcite{r17}, which  performed meta-learning on a pseudo decision boundary between known and outlier points. Some methods used the generative models to create pseudo-unknown samples to bolster the classifier. For example, Pal et al.\coloredcite{r51} used two GANs to generate samples for both closed and open spaces, enhancing the discriminative model to predict unknown class samples. At the same time, some methods combined the discriminative and generative methods. For example, Sun et al.\coloredcite{r14} introduced representative point learning (RPL) to model out-of-class space using known classes, employing a learnable dynamic threshold strategy for each known classes to reject unknown classes.

While addressing the open-set HSI classification challenge, current methods focus mainly on distinguishing the unknown class samples from the known class samples and rejecting them to increase the accuracy of identifying known class samples. They fails to further identify or discovery the unknow classes. Essentially, unknown class discovery extends unknown class rejection by requiring the model not only to reject unknown samples but also to further classify different unknown classes. In other words, unknown class discovery is a clustering task, where the model should transfer the knowledge learned from labeled known classes to cluster the unknown classes in unlabeled data, which can be referred to as deep transfer clustering\coloredcite{r40}. To address the discovery of unknown classes, some methods have been designed in the computer vision domain by now. When mainly focusing on the natural images, these methods primarily used pseudo-labels to transfer knowledge from labeled known class samples to unlabeled unknown class samples\coloredcite{r41},\coloredcite{r42},\coloredcite{r10},\coloredcite{r23},\coloredcite{r53}. For example, Han et al.\coloredcite{r42} utilized similarity pairs with custom thresholds for pseudo-labeling to transfer the knowledge of known classes to discover unknown classes; Cao et al.\coloredcite{r10} employed the cosine distance between feature representations and an uncertainty-adaptive margin mechanism to generate high-quality pseudo-labels for unlabeled samples, with pairwise loss grouping similar unlabeled samples together. Nevertheless, these methods have not addressed the unknown class discovery under the few-shot setting where are only few labeled samples available. Thus, it is still challenging to the discovery of unknown classes in HSIs\coloredcite{r15}. 

In this paper, we propose a prototype learning and clustering method for discoverying unknown classes in HSIs under the few-shot environment. Using few labeled samples, our idea is to learn to distinguish unknown classes  from known classes, and discovery the potential unknown classes by infering the prototypes of unknown classes. In testing stage, the samples will be first classified as known or unknown classes with the learned classifier, and the unknow class samples will be further clustered into different classes according to their distance to the inferred unknown class prototypes. To distinguish unknown classes from known classes, a class anchor based classification strategy is adopted, which expands the original dimensional logit space to include an additional dimension specifically for representing unknown class. This enables the classifier to recognize the unknow class samples while classifying these known class samples. Meanwhile, when discoverying the potential unknow classes, a prototype clustering method is used, which initializes multiple trainable prototypes whose number is significantly greater than that of actual unknown classes and then clusters the prototypes into different groups representing the potential unknown classes. The main contributions of our work are summarized as follows:

\begin{enumerate}
	\item We perform known class classification and detect unknown class samples through class anchor based open-set classification.
	\item Building upon the detected unknown class samples, we cluster distinct unknown classes and estimate their quantity using a dual-level prototype contrastive learning module.
	\item We conducted a comprehensive experimental evaluation. The results demonstrate that the proposed method for unknown class discovery in HSIs significantly outperforms state-of-the-art approaches. 
\end{enumerate}

The rest of this paper is organized as follows. 
Section \ref{H} introduces the proposed few-shot learning method for HSI unknown class discovery.
Section \ref{H1} shows experimental results
to demonstrate the superior performance of our method. Finally, 
Section \ref{H2} concludes this article.

\section{PROPOSED METHODOLOGY}\label{H}
In this section, we state the problem of few-shot HSI unknown class discovery and detail the method proposed.

\subsection{Problem statement}
The fundamental challenge addressed by few-shot unknown class discovery \coloredcite{r33},\coloredcite{r34} is to accurately identify and classify previously unseen classes using only a limited number of training samples from known classes. 

In the context of few-shot learning, there is one target dataset $D_t$ which encompasses $C$ class samples and only few samples per class are labeled (i.e., 5 samples per class). Under the assumption that all the unlabeled samples belong to the same classes of these few labeled samples, the problem of few-shot learning is to infer the classes of these unlabeled samples by using the few labeled samples.  To address the problem, the $C-way$ $K-shot$ tasks are often constructed in the episodes of model training process by following the meta-learning, aiming at equiping the model with the ability to generalize under conditions of changing classes. By augmenting these few-shot labeled samples or using a source dataset which consists of a large number of labeled samples, $C$ classes and $(K+M)$ smaples per class are selected each time from the labeled samples to compose a task. These selected samples will consist of a support set and a query set. There are $K$ samples per class, i.e., $C\times K$ samples in the support set denoted as $\left\{\left(x_i, y_i\right)\right\}_{i=1}^{C \times K}$, while there are $C\times M$ samples in the query set denoted as $\left\{\left(x_i, y_i\right)\right\}_{i=1}^{C \times M}$.  The training purpose is to enable the model to correctly predict the classes of the samples in the query set according to their distance to the labeled samples in the support set. In the testing phase, all the samples in the query set are unlabeled, whose classes will be predicted by using the trained model in the same way as that of traing stage. 

The few-shot unknown class discovery problem addressed in this paper is more challenging than the few-shot learning problem described above. That is, the accurate  number of classes in $D_t$ is unknown, or some of the $C$ classes in $D_t$ lack the labeled samples. This leads to that, these unlabeled samples in the query set may be not all from the known classes of these labeled samples in the support set. Our purpose in this paper is to infer the classes of these query samples from the known classes, but also discovery the potential unknown classes among these query samples. Therefore, given the $C$ classes in $D_t$, we suppose that there are $C_k$ known classes and $C_u$ unknown classes, where $C=C_k \cup C_u$. And when constructing the tasks of few-shot learning, while all the samples of support set are from the $C_k$ known classes, some samples of query set will be from the $C_k$ known classes and the rest will be from the  $C_u$ unknown classes.

\subsection{Model Overview}

To resolve the few-shot HSI unknown class discovery problem stated above, this paper proposes a prototype learning and clustering method. The model architecture is shown in Fig. \ref{H31}. Given one task composed by a support set and a query set, the proposed method utilizes a deep 3D residual network as the feature extractor to obtain spatial-spectral embedding features. Following that, there are two learning steps: open-set classification and unknown class discovery. 

\textbf{Open-set Classification}: The open-set classification step employes a class anchor based classifier that classifies the samples from known classes and rejects these from unknown classes. Representing the prototypes of each class, the class anchors $\mathcal{A}=\left(\mathbf{a}_1,\ldots, \mathbf{a}_{k}, \mathbf{a}_{k+1}\right)$ include $k$ known class anchors and one unknown class anchor. By classifying all the samples in the task to these anchors, there are two kinds of losses generated in training process, i.e.,  $L_{osc}$ and $L_{ca}$ showing in Fig. \ref{H31}. The former is the open-set classification loss, while the latter is the loss enforcing the samples closer to their true class anchors and further from other anchors. 

\textbf{Unknown Class Discovery}: The unknown class discovery step utilizes a dual-level semi-supervised prototype contrastive learning strategy to explore the potential classes.  It introduces multiple trainable prototypes $\mathcal{C}=\left\{\mathbf{c}_1,\ldots,\mathbf{c}_{w}\right\}$, where the number of prototypes $w$ is predefined to be significantly greater than the number of actual  classes. For each sample in the input task, two agumented samples named weakly augmented sample $x$ and strongly augmented sample  $x^{\prime}$ are generated with random augmentations (cropping, coloring, resizing) to increase the diversity of samples, instead of using the input samples directly.  Given the augmented samples and predefined prototypes, there are four computation steps followed by different losses.

\begin{enumerate}
	\item[*]\textbf{Prototype Prediction}: The proposed method first tries to assign each sample to these prototypes, and generates a prototype similarity loss $L_{ps}$ by constraining the samples in each positive sample pair exhibiting similar  probability distribution. From the perspective of the weakly augmented samples, the positive sample of each labeled weakly augmented sample is randomly selected from these labeled strongly augmented samples with the same class, and the positive sample of each unlabeled weakly augmented sample is the unlabeled strongly augmented sample with highest cosine similarity.
	
	\item[*]\textbf{Prototype Group Prediction}: Once the samples are assigned to different prototypes and the probabilies are calculated, the similarities between each prototypes are calculated based on Jaccard distance. And according to the similarities, the prototypes are merged into groups based on Louvain clustering algorithm\coloredcite{r21}. Then, the probabilities that each sample belongs to different prototype groups are further computed, and accordingly the prototype group similarity loss $L_{pgs}$ is generated by requiring the samples in the positive sampe pairs to share similar group probability distribution. 
	
	\item[*] \textbf{Prototype Regularization}: The prototype regularization, i.e., $L_{reg}$ , is also employed to avoid neglecting prototypes or groups with lower probability distributions under class imbalance, which may lead all samples to be assigned to the same prototype.	
	
	\item[*]\textbf{Known Class Prediction}: Known class prediction step calculates the loss $L_{kcd}$  to maintain and enhance the classification capability for known classes. 
\end{enumerate}

Based on above steps, the proposed method progressively clusters prototypes into prototype groups.  The overall loss function of the proposed method is defined as shown in Eq.\eqref{eq:total}, where these losses in the function will be detailed in the following subsection $D$ and $E$. Particularly, before the training process, there is  a pre-training process which just executes the open-set classification step to learn the class anchors. The pre-training process is described in Algorithm \ref{alg:FSOSR2}, while the training process is shown in Algorithm \ref{alg:FSOSR}.

\begin{equation}
L_{total}=L_{class}+L_{disc}
\label{eq:total}
\end{equation}

\begin{equation}
L_{class}=L_{osc}+L_{ca}
\label{eq:class}
\end{equation}

\begin{equation}
L_{disc}=L_{\text {ps}}+L_{\text {pgs}}+L_{\text {reg}}+L_{kcd}
\label{eq:disc}
\end{equation}

\begin{figure*}[htbp]
	\centering
	\graphicspath{picture_3.pdf}
	\caption{The architecture of the proposed model. Once a few-shot learning task consisting of a support set and a query set, the spatial-spectral features of each sample will be extracted by a deep 3D residual network. Then, two learning steps are followed. First, open-set classification step is to classify the samples from known classes and reject the samples from unknown classes. Initializing a set of $N$ anchors representing $N-1$ known classes and all the unknown classes, the training process is to enforce the samples close to their true anchors.  Second, unknow class discovery step is to discovery the potential different classes among the samples by introducing multiple trainable prototypes and clustering the prototypes progressively, where the prototype groups represent various classes. }	
	\label{H31}
\end{figure*}

\textbf{The testing process}: As shown in  Algorithm \ref{alg:FSOSR1}, during the testing phase, the open-set classifier is first applied to identify samples of unknown classes and infer the classes of these samples from known classes. For those samples identified as belonging to unknown classes, their unknown classes affiliations will be further inferred by  performing fine-grained clustering based on pairwise similarity measures of prototypes and prototype groups.

Next, we provide a detailed description of the main components of the proposed method as follows.

\begin{table}[h]
	\small
	\caption{The architecture of the feature extractor}
	\label{model-modules}
	\centering
	\renewcommand{\arraystretch}{1.5}
	\setlength{\tabcolsep}{2.7mm}{
		\begin{tabular}{c|c||c}
			\hline
			\multicolumn{2}{c||}{Model Architecture} & SIZE \\ \hline\hline
			\multicolumn{2}{c||}{INPUT} & 9$\times$9$\times d$ \\ \hline\hline
			\multirow{1}{*}{Convolutional layer}& Conv2d & 1×1(100) \\
			\multirow{1}{*}{}& BatchNorm2d + Relu & N/A \\ \hline
			\multirow{8}{*}{Residual block1} 
			& Conv3d & 3×3×3(8) \\
			& BatchNorm3d + Relu & N/A \\
			& Conv3d & 3×3×3(8) \\
			& BatchNorm3d + Relu & N/A \\
			& Conv3d & 3×3×3(8) \\
			& BatchNorm3d + Relu & N/A \\ \hline
			\multirow{1}{*}{MaxPool layer} & MaxPool3d & 4×2×2   \\ \hline    
			\multirow{8}{*}{Residual block2} 
			& Conv3d & 3×3×3(16) \\
			& BatchNorm3d + Relu & N/A \\
			& Conv3d & 3×3×3(16) \\
			& BatchNorm3d + Relu & N/A \\
			& Conv3d & 3×3×3(16) \\
			& BatchNorm3d + Relu & N/A \\ \hline
			\multirow{1}{*}{MaxPool layer} & MaxPool3d & 4×2×2   \\ \hline     
			\multirow{1}{*}{Convolutional layer} &  Conv3d  & 3×3×3(32) \\ \hline
			\multirow{1}{*}{Fully Connected layer} &   & 160×$N$ \\ \hline
			\multirow{1}{*}{OUTPUT} & & 1$\times$$N$ \\
			\hline\hline
		\end{tabular}
	}
\end{table}

\subsection{Feature Extractor}
The architecture of the proposed feature extractor $f_\theta$ for HSI spectral-spatial information extraction is illustrated in Table \ref{model-modules}. For each spatial pixel location in HSI data, a three-dimensional spectral-spatial patch $I \in \mathbb{R}^{H \times W \times B}$ is formed, where $H$, $W$, and $B$ denote height, width, and spectral bands, respectively. The backbone of  $f_\theta$ is based on a 3D CNN architecture, primarily composed of 3D residual convolution layers, which are capable of learning enhanced spectral-spatial features. Each residual block consists of three consecutive Conv3D units. Following each residual block, a 3D max-pooling layer reduces computational complexity and aggregates spectral-spatial relationships, ultimately forming the spectral-spatial embedding feature $\mathcal{Z}$.

\subsection{   Class Anchor based Open-set Classification}
Given a task consisting of the samples from known and unknown classes, open-set classification is to classify these samples from known classes and reject the samples from unknown classes. To achieve this goal, we applied a class-anchor based approach. That is, we use $N$ anchors to represent the prototypes of these classes, where $N-1$ anchors representing the known classes and the remaining one anchor representing the unknown classes. And then, we train the classifier by enforcing the samples close to their true anchors. 

Each anchor is initialized with one-hot vector which is adjusted by scaling parameters $\varphi$ to increase class variance. This can be described as follows:

\begin{equation}
\begin{gathered} 
\mathcal{A}=\left(\mathbf{a}_1, \ldots, \mathbf{a}_{N}\right)=\left(\varphi \cdot \mathbf{e}_1, \ldots, \varphi \cdot \mathbf{e}_{N}\right) \\
=\left((\varphi, 0, \ldots, 0)^{\top}, \ldots,(0,0, \ldots, \varphi)^{\top}\right)
\label{eq:4}
\end{gathered}
\end{equation}

To assign the samples to these anchors, the distance between the spatial-spectral embedded feature $\mathcal{Z}$ of each sample and various class anchors $\mathbf{d}=\left(\mathbf{d}_1, \ldots, \mathbf{d}_{N}\right)=\left(\left\|\mathbf{z}-\mathbf{a}_1\right\|_2, \ldots,\left\|\mathbf{z}-\mathbf{a}_{N}\right\|_2\right)^{\top}$ are established, where $\|\cdot \|_2$  denotes the Euclidean norm. 

For the samples in the input task, each sample will be first augmented with two new samples by random augmentations, i.e., weakly augmented sample $x$ and strongly augmented sample  $x^{\prime}$. This means when there are $M$ samples in the task, there will be $2M$ samples augmented, which will be used for traing in one episode. Then, with these initialized anchors, the cross-entropy based open-set classification loss $L_{osc}$ as shown in Eq.\eqref{eq:osc} is generated in training process, where $y_i$ is the label of $i$-th sample. It is commonly understood that cross-entropy loss minimizes the negative log probability of the true classes for the input, achieved through the normalization of the logarithm using the softmax function. The softmax function is not injective; multiple input logit vectors can map to the same output softmax probability vector \coloredcite{r16}, which complicates the ability of classifiers to identify and reject unknown classes. To address this, we expand the original $N-1$ dimensional logit space to include
an additional dimension specifically for representing unknown
class. And a pseudo-label is used for the samples from all the unknown classes during training.  

\begin{equation}
L_{osc}=-\frac{1}{2M} \sum_{i=1}^{2M} \log \left(\frac{e^{-\mathbf{d}_{y_{i}}}}{\sum_{j=1}^N e^{-\mathrm{d}_j}}\right)
\label{eq:osc}
\end{equation}

Moreover, to enhance the model's classification performance, a class anchor loss $L_{ca}$ is also used to  encourage the training samples to be as close as possible to their corresponding true class anchors while distancing them from all other class anchors, which can be seen in Eq. \eqref{eq:ca}.  It aims to strengthen the model's discriminative power by minimizing the distance between the input samples and their true class anchors, and maximizing the distance to other class anchors. There is a hyper-parameter  $\gamma$ to adjust the penalty item which is applied on the Euclidean distance between the features of the input samples and their true class anchors. $d_y$ denotes the distance of the sample to its ture anchor, and $z_i$ is its spatial-spectral feature produced by the feature extractor. 

\begin{equation}
L_{ca}=\frac{1}{2M} \sum_{i=1}^{2M} \log \left(1+\sum_{j \neq i}^{N} e^{d_y-d_j}\right)+\gamma\left\|\mathbf{z}_i-\mathbf{a}_y\right\|_2)
\label{eq:ca}
\end{equation}

During the pre-training stage, the class anchors corresponding to known classes are dynamically updated using labeled support set samples through the adaptive mechanism formulated in Eq. \eqref{eq:ab}.  Let $B_s$ be the set of samples for known class $B$, $\mathbf{a}_B$ is adjusted according to the average Euclidean distance from the sample of the class $B$ to the class anchor. The distance is calculated by following the principle that the closer a sample is to the class anchor, the higher the probability it belongs to that class. Therefore, we employ the softmin function in Eq. \eqref{eq:ab}, which is the inverse of the softmax function.

\begin{equation}
\mathbf{a}_B=\frac{1}{\left|B_s\right|} \sum_{i=1}^{B_s} \mathbf{d}_i \circ\left(1-\operatorname{softmin}\left(\mathbf{d}_i\right)\right)
\label{eq:ab}
\end{equation}

\subsection{Prototype Learning and Clustering based Unknow Class Discovery}
While classifying the samples from known classes and rejecting the samples from unknown classes, the step of open-set classification can only seperate the unknown class samples from these of known classes, without discoverying the potential unknown classes. Different from the open-set classification, the unknow class discovery is to discovery the potential different classes among the samples, including the known and unknown classes, by following a prototype learning and clustering based approach. As mentioned, it introduces multiple trainable prototypes $\mathcal{C}=\left\{\mathbf{c}_1,\ldots,\mathbf{c}_{w}\right\}$, where the number of prototypes $w$ is significantly greater than the number of actual  classes. Through  assigning each samples to each prototype and then clustering the prototypes during training, the model learns to discovery the potential sample classes which are represented by these prototype groups.

Specially, given the weakly augmented samples $x$ and strongly augmented samples  $x^{\prime}$, the positive sample pairs are first built for each sample. For each weakly augmented sample $x$ from support set where the labels are available, its positive sample is randomly selected from these strongly augmented sample  $x^{\prime}$ within the same class. Meanwhile, for the weakly augmented samples $x$ from query set where there are none labels, the positive samples are selected from these strongly augmented samples according to the cosine similarity.  

With these built positive sample pairs and the feature representations of each sample, the probability of one sample belonging to $k$-th prototyp is calculated by Eq. \eqref{eq:8}. 

\begin{equation}
\mathbf{p}^{(k)}=\frac{\exp \left(g\left(\mathbf{z}, \mathbf{c}_k\right) / \tau\right)}{\sum_{c_{k^{\prime}} \in C} \exp \left(g\left(\mathbf{z}, \mathbf{c}_{k^{\prime}}\right) / \tau\right)}
\label{eq:8}
\end{equation}

In this context, $\tau$ represents a temperature parameter that controls the smoothness of the distribution ($\tau=0.1$ in our experiments), $\mathbf{c}_k$ refers to the $k$-th prototype, and $g\left(\mathbf{z}, \mathbf{c}_k\right)$ denotes the dot product between $\mathbf{c}_k$ and the embedding feature $\mathbf{z}$ of the sample.

Then, to constrain the samples in the positive sample pairs to share similar probabilities belonging to each prototype, the prototype similarity loss $L_{ps}$is calculated, which is defined as Eq. \eqref{eq:ps}. 

\begin{equation}
L_{\text {ps}}=-\frac{1}{M} \sum_i^{M} \log \left(\mathbf{p}_i \cdot \mathbf{p}_i^{\prime}\right)
\label{eq:ps}
\end{equation}

Here, $\mathbf{p}_i$ denotes the probability of a sample with respect to various prototypes, while $\mathbf{p}_i^{\prime}$ represents the probability of its positive sample relative to the various prototypes.


Given the derived prototype groups, it is expected that the samples in the positive sample pairs also share the similar probabilities of belonging to different prototype groups. Accordingly, the prototype group similarity loss $L_{pgs}$ is calculated as Eq.\eqref{eq:pgs}.  

\begin{equation}
L_{\text {pgs}}=-\frac{1}{M} \sum_i^{M}\left(\mathbf{q}_i^{\prime} \log \mathbf{q}_i+\mathbf{q}_i \log \mathbf{q}_i^{\prime}\right)
\label{eq:pgs}
\end{equation}

$\mathbf{q}_i$ represents the probabilities of $i$-th sample to each prototype group, while $\mathbf{q}_i^{\prime}$ denotes the probabilities of its positive sample to each prototype group. The values of  $\mathbf{q}_i$ and $\mathbf{q}_i^{\prime}$  are computed according to Eq. \eqref{eq:12}, where $\mathcal{C}_g$ is defined as a set of prototypes in one prototype group.  

\begin{equation}
\mathbf{q}_i=\frac{\sum_{\mathbf{c}_k \in \mathcal{C}_g} \exp \left(g\left(\mathbf{z}, \mathbf{c}_k\right) / \tau\right)}{\sum_{\mathbf{c}_{k^{\prime}} \in \mathcal{C}} \exp \left(g\left(\mathbf{z}, \mathbf{c}_{k^{\prime}}\right) / \tau\right)}
\label{eq:12}
\end{equation}

Besides above losses, there are two more loss functions which are also used during the training. First, to avoid that many samples are assigned to one same prototype, a prototype regularization loss $L_{reg}$ is adopted to allow the probability $P_i$ of a sample with respect to various prototypes approaches the prior probability distribution, i.e., a balanced distribution among different prototypes, $\mathbf{p}_{\text {prior }}^{(k)}=\frac{1}{\mathcal{N}_g \times\left|\mathcal{C}_k\right|}$.  $\mathcal{N}_g$ represents the total number of prototype groups in the current phase, and $\left|\mathcal{C}_k\right|$ signifies the number of prototypes in the group belonging to $\mathcal{C}_k$.

\begin{equation}
L_{\text {reg }}=K L\left(\frac{1}{M} \sum_i^{M} \mathbf{p}_i \| \mathbf{p}_{\text {prior }}^{(k)}\right)
\label{eq:reg}
\end{equation}

Second, to avoid the risk of incorrectly assigning unknown class samples to the known class prototypes, we further use the Hungarian algorithm to match known class labels with prototype groups, and calculate the known class discovery loss $L_{kcd}$ based on these labeled support samples.

\begin{equation}
L_{kcd}=-\frac{1}{D} \sum_i^D \left(\log \mathbf{q}_i^{(y)}+\log \mathbf{q}_i^{\prime(y)}\right)
\label{eq:kcd}
\end{equation}

$D$ is the number of support samples which are from known classes. $\mathbf{q}_i^{(y)}$ and $\mathbf{q}_i^{\prime(y)}$ respectively represent the probabilities of correct alignment with the prototype groups corresponding to the actual labels, while $y_i$ is the actual label of the $i$-th support sample.

In the prototype updating stage,once the probabilies of each sample belonging to different prototypes are predicted, the prototypes will be further clustered into prototype groups. The samples assigned to the same prototypes are highly likely to belong to the same class, and conversely, the more identical samples there are between two prototypes, the more likely it is that these prototypes belong to the same prototype group. Therefore, by assigning the top three prototypes with the highest allocation probabilities for each sample as the associated prototypes and using a similarity matrix based on Jaccard distance to estimate the similarity between prototypes, the similarity between prototype $i$ and prototype $j$ can be calculated by following Eq. \eqref{eq:10}.

\begin{equation}
	s_{i j}=\frac{\left|\Gamma\left(\mathbf{c}_i\right) \cap \Gamma\left(\mathbf{c}_j\right)\right|}{\left|\Gamma\left(\mathbf{c}_i\right) \cup \Gamma\left(\mathbf{c}_j\right)\right|}
	\label{eq:10}
\end{equation}

where $\Gamma\left(\mathbf{c}_i\right)$ represents the sample set of $i$-th prototype. With the probabilities between different prototypes.Then, based on the similarity matrix between prototypes, the model uses the Louvain clustering algorithm\coloredcite{r21} to cluster hierarchically to form a dynamic prototype group, and then establishes a bidirectional mapping relationship between the prototype group of known categories and the real class labels by Hungarian Algorithm\coloredcite{r11}, and the remaining prototype groups are the discovered unknown classes.
\begin{algorithm}
	\caption{Pre-Training Steps}
	\label{alg:FSOSR2}
	\begin{algorithmic}[1]
		\renewcommand{\algorithmicrequire}{\textbf{Input:}} 
		\REQUIRE feature extractor $f_\theta$, pretrain set  $\mathcal{D}_{pre}$, class anchors $\mathcal{A}$
		\renewcommand{\algorithmicensure}{\textbf{Output:}}
		\ENSURE Updated $f_\theta$, $\mathcal{A}$ 
		\STATE Extract features: $\mathcal{Z}_{pre} \leftarrow f_\theta(\mathcal{D}_{pre})$ 
		\STATE Update $f_\theta$ by optimizing the open-set classification loss $L_{class}$ shown in Eq. \eqref{eq:class};
		\STATE Update $\mathcal{A}$ through Equation \eqref{eq:ab};
		\RETURN Updated parameters of $f_\theta$, $\mathcal{A}$;
	\end{algorithmic}
\end{algorithm}

\begin{algorithm}
	\caption{Training Steps}
	\label{alg:FSOSR}
	\begin{algorithmic}[1]
		\renewcommand{\algorithmicrequire}{\textbf{Input:}} 
		\REQUIRE feature extractor $f_\theta$, support set, query set, trainable prototypes $\mathcal{C}$, prototype groups $\mathcal{Q}$
		\renewcommand{\algorithmicensure}{\textbf{Output:}}
		\ENSURE Updated $f_\theta$, $\mathcal{C}$ parameters, updated $\mathcal{Q}$
		\STATE Augment the samples in the support set and query set and built the positive sample pairs $<x,x^{\prime}>$;
		\STATE Extract features through $f_\theta$;
		\STATE Using all the augmented samples to calculate the open-set classification loss $L_{osc}$ shown in Eq.\eqref{eq:osc} and the class anchor loss $L_{ca}$ shown in Eq. \eqref{eq:ca};
		\STATE Compute the probabilities that each sample belongs to different prototypes and derive the prototype similarity loss $L_{ps}$ shown in Eq. \eqref{eq:ps};
		\STATE Compute the similarities between different prototypes and the prototype group similarity loss $L_{pgs}$ shown in Eq.\eqref{eq:pgs};
		\vspace{-2.5ex} 
		\STATE Calculate the prototype regularization loss $L_{reg}$ shown in Eq.\eqref{eq:reg} and the known class discovery loss $L_{kcd}$ shown in Eq.\eqref{eq:kcd};
		\STATE Update $\mathcal{Q}$ by applying the Louvain clustering algorithm to prototypes to group them;
		\STATE Update $f_\theta$, $\mathcal{C}$ with the losses;
		\RETURN Updated parameters of $f_\theta$ and $\mathcal{C}$, updated prototype groups $\mathcal{Q}$.
	\end{algorithmic}
\end{algorithm}    
\begin{algorithm}
	\caption{Testing Steps}
	\label{alg:FSOSR1}
	\begin{algorithmic}[1]
		\renewcommand{\algorithmicrequire}{\textbf{Input:}} 
		\REQUIRE feature extractor $f_\theta$, test set $\mathcal{D}_{test}$, class anchors $\mathcal{A}$, trainable prototypes $\mathcal{C}$, prototype groups $\mathcal{Q}$
		\renewcommand{\algorithmicensure}{\textbf{Output:}}
		\ENSURE Classification of the test samples
		\STATE Extract features:$\mathcal{Z}_{test} \leftarrow f_\theta(\mathcal{D}_{test}) $
		
		\IF { The sample is closest to the unknown class anchor $\mathcal{A}_{N}$ }
		\STATE Classify the sample as unknown class;
		\STATE Assign the sample to the prototype group representing unknown classes;
		\ELSE
		\STATE Predict the label of the sample based on its distance to the different anchors of known classes shown in \eqref{eq:4};
		\ENDIF
		\RETURN Predicted classes of the test samples;
	\end{algorithmic}
\end{algorithm}

\section{EXPERIMENT}\label{H1}
\subsection{Hyperspectral Datasets Description} 
To fairly assess the performance of the proposed methods, we have selected datasets that are widely used in related work for training and testing. Specifically, we evaluate our methods on the IP, PU, WHU-Hi-HanChuan and SA datasets. We provide a brief introduction to these datasets as follows.

\begin{figure}[htbp]
	\centering
	\subfigure[]{
		\includegraphics[width=3cm,height=3cm]{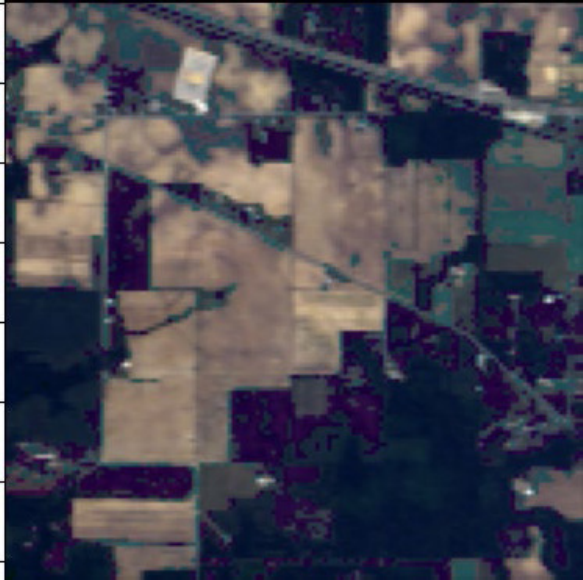}
	}
	\subfigure[]{
		\includegraphics[width=3cm,height=3cm]{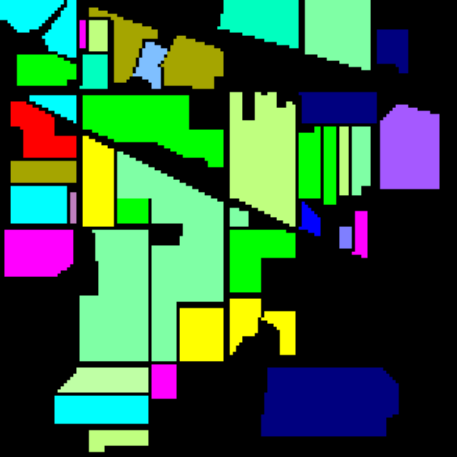}
	}
	\subfigure[]{
		\includegraphics[width=1.5cm,height=3cm]{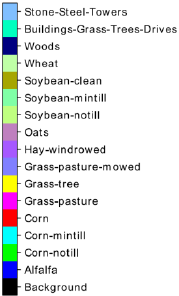}
	}
	
	\caption{ IP dataset color map, label map, and label color}\label{H6}
\end{figure}

(a) \textbf{IP dataset}: It was imaged by an airborne visible infrared imaging spectrometer (AVIRIS) in Indiana, USA. This data contains 200 bands with the wavelength ranging from $0.4-2.5m$. The size is $145\times145$ pixels, with a spatial resolution of about 20m. As shown in Fig.\ref{H6}, there are 16 classes of land cover  including crops and natural vegetation. Among these classes,  11 classes  were selected as known classes in our experiments, while the remaining 5 classes were treated as unknown classes.

\begin{figure}[htbp]
	\centering
	\subfigure[]{
		\includegraphics[width=1.5cm,height=3cm]{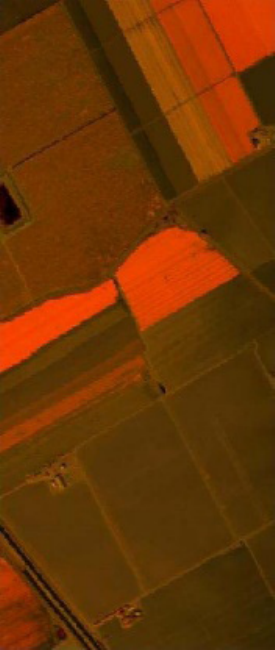}
	}
	\subfigure[]{
		\includegraphics[width=1.5cm,height=3cm]{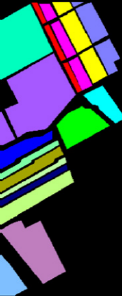}
	}
	\subfigure[]{
		\includegraphics[width=3.5cm,height=3cm]{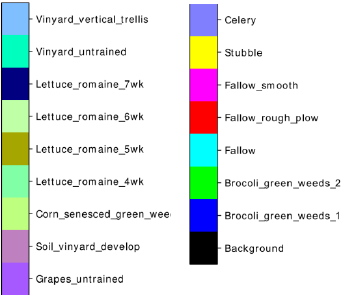}
	}
	
	\caption{SA dataset color map, label map, label color}\label{H7}
\end{figure}

(b)\textbf{ SA dataset}: This dataset was captured  by the airborne visible infrared imaging spectrometer (AVIRIS) in the Salinas Valley, California, USA. It contains 204 wavebands with a size of $512 \times 217$, and a spatial resolution of about $3.7m$. As shown in Fig.\ref{H7}, there are 16 classes of land cover, which includes  vegetables, exposed soil, etc. Similarly, for the SA dataset, the first 11 classes were selected as known classes, while the remaining 5 were used as unknown classes in our experiments.

(c) \textbf{PU dataset}: This dataset was imaged by the German airborne reflective optical spectral imager (ROSIS) in the city of Pavia, Italy. The data contains 103 wavebands, with a size of $610 \times 340$ pixels and a spatial resolution of about $1.3m$. As shown in Fig. \href{H8}{4}, there are 9 classes of land cover, including trees, asphalt roads, bricks, etc. Also for experiments, 5 classes of PU dataset were selected as known classes, and the rest of 4 classes were taken as unknown classes. More information about these datasets can be found from \url{https://www.ehu.eus/ccwintco/index.php/Hyperspectral_Remote_Sensing_Scenes}.

\begin{figure}[htbp]
	\centering
	\subfigure[]{
		\includegraphics[width=2cm,height=3cm]{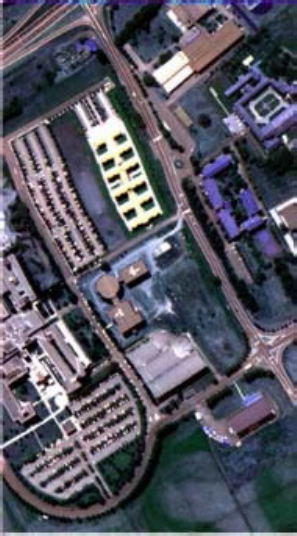}
	}
	\subfigure[]{
		\includegraphics[width=2cm,height=3cm]{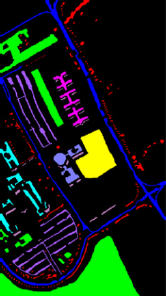}
	}
	\subfigure[]{
		\includegraphics[width=1.2cm,height=3cm]{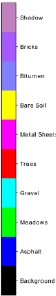}
	}
	
	\caption{ PU dataset color map, label map, label color}\label{H8}
\end{figure}

(d) \textbf{WHU-Hi-HanChuan dataset}\coloredcite{r67},\coloredcite{r68}: This dataset was acquired by Headwall Nano-Hyperspec imaging sensor equipped on a Leica Aibot X6 UAV platform in Hanchuan, Hubei province, China. There are 1217 × 303 pixels, 274 bands from 400 to 1000 nm, and the spatial resolution is about 0.109 m. As shown in Fig. \href{H9}{5}, there are 16 classes of land covers, including strawberry, cowpea, soybean, sorghum, water spinach, etc. Also for experiments, 11 classes of WHU-Hi-HanChuan dataset were selected as known classes, and the rest of 5 classes were taken as unknown classes. More information about the IP, SA and PU datasets can be found from \url{https://www.ehu.eus/ccwintco/index.php/Hyperspectral_Remote_Sensing_Scenes} and WHU-Hi-HanChuan dataset is available at \url{http://rsidea.whu.edu.cn/resource\_WHUHi\_sharing.htm}.

\begin{figure}[htbp]
	\centering
	\subfigure[]{
		\includegraphics[width=2cm,height=3cm]{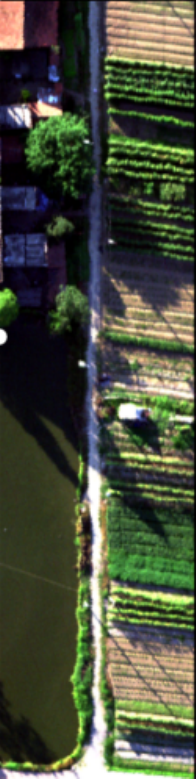}
	}
	\subfigure[]{
		\includegraphics[width=2cm,height=3cm]{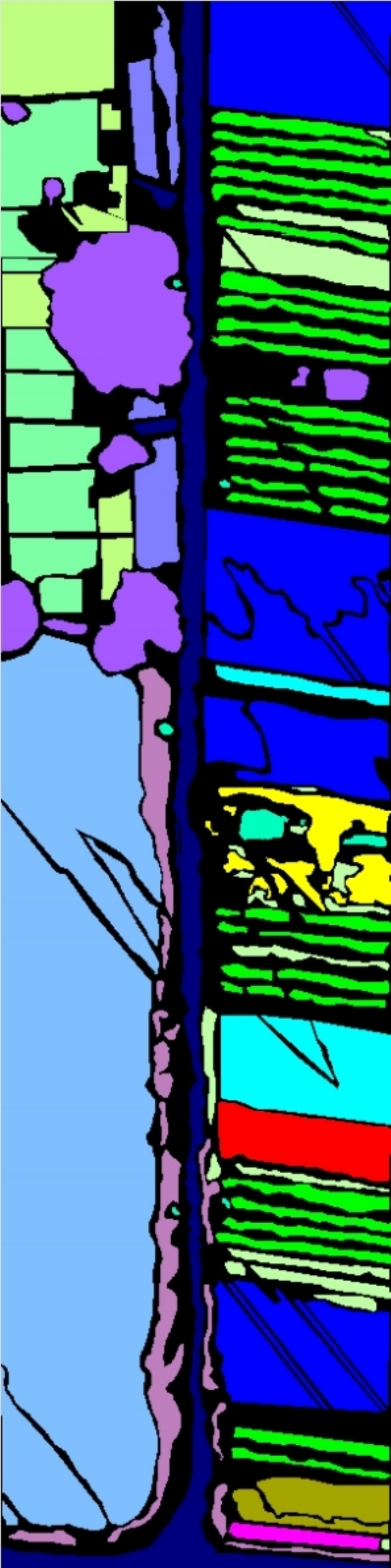}
	}
	\subfigure[]{
		\includegraphics[width=1.5cm,height=3cm]{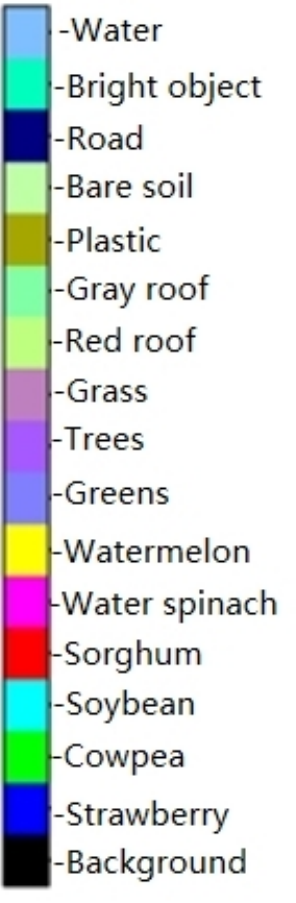}
	}
	
	\caption{ WHU-Hi-HanChuan dataset color map, label map, label color}\label{H9}
\end{figure}
\subsection{Experimental Setup}
\textbf{Evaluation Metrics:} Since the discovery of new classes in few-shot HSIs is fundamentally  a deep transfer clustering task, we adhere to the evaluation metrics described in \coloredcite{r10}, assessing the performance of our model on both known and unknown classes. In detail, there are three metrics used for evluation.

\begin{enumerate}
	\item[*]\textbf{Known ACC}: This metric refers to the classification accuracy of known classes, measuring the percentage of seen class samples correctly classified by the model.
	
	\item[*]\textbf{Unknown ACC}: Unknown ACC measures the classification accuracy of unknown classes. As our approach identifies unknown class samples through clustering, clustering accuracy is utilized for assessment and computed by solving the predicted target class assignment using the Hungarian algorithm\coloredcite{r11}. 
	
	\item[*] \textbf{ALL ACC}: ALL ACC calculates the clustering accuracy for all known and unknown class samples to measure the overall performance of the proposed model.
	
\end{enumerate}

\textbf{Implementation Details:} Our approach employs an episodic learning strategy to train a deep residual 3-D CNN to extract discriminative features from samples. In the training phase,  $k+d$ samples per class for known classes and $d$ samples per class for unknown classes are randomly selected to form a task in each episode.  Among these selected samples,  $k$ samples per class from known classes will form the support set, while all the rest of samples from known and unknown classes will form the query set. In our experiments, we set $d=3k$ and $k=1$ or $5$. Due to the scarcity of hyperspectral samples, the pre-training set employs data augmentation techniques, incorporating random Gaussian noise into existing training samples to expand the pre-training dataset. As described in references\coloredcite{r12} and\coloredcite{r13}, this approach has been widely adopted. Moreover, there are 35 prototypes initialized on SA and IP datasets, while 25 prototypes are initialized on PU dataset, while 40 prototypes are initialized on WHU-Hi-HanChuan dataset. The proposed method operates efficiently on a computer equipped with an Intel Core i9-9900K CPU and NVIDIA GeForce RTX 3090 GPU, requiring 64GB RAM and 25.4GB storage space.

\subsection{Performance Comparison of Methods}
To evaluate the proposed approach,  we selected state-of-the-art (SOTA) methods as baselines for comparision. They are  SSMLP-RPL\coloredcite{r14}, MDL4OW\coloredcite{r15}, DFSL\coloredcite{r16}, OCN\coloredcite{r17}, MORGAN\coloredcite{r18}, and EVML\coloredcite{r19}. Because most baseline methods fail to estimate the number of unknown classes without prior knowledge, we do the comparision  under the assumption that the aforementioned methods are aware of the number of unknown classes.

For DFSL\coloredcite{r16}, it lacks the ability to discovery new classes mixed in the unlabeled data under open settings. Therefore, we introduced an additional softmax function into the network and set a threshold of 0.5 to recognize unknown classes. When the output with the highest probability is less than 0.5, the samples were considered as unknown classes. SSMLP-RPL\coloredcite{r14} constructs an out-of-class space using reciprocal points, pushing known class samples towards the edge of the space and distinguishing unknown classes within the space via a learnable dynamic threshold. OCN\coloredcite{r17} and MORGAN\coloredcite{r18} train outlier detectors based on the Euclidean distance from test queries to known class prototypes to reject unknown class samples. EVML\coloredcite{r19} employs a P-OpenMax layer to reject unknown class samples, differentiating unmarked query samples based on a Weibull model fitted to the distance from support features to their respective prototypes. 

SSMLP-RPL\coloredcite{r14}, MDL4OW\coloredcite{r15}, OCN\coloredcite{r17}, and MORGAN\coloredcite{r18} can recognize and reject unknown class samples through empirical thresholds or anomaly detectors, while EVML\coloredcite{r19} is able to identify unknown classes as one coarse class without further discoverying different unknown classes in the unknown class samples. Hence, to facilitate the comparision, we further employ the k-means algorithm to cluster the unknown class samples recognized by these methods to discover distinct unknown classes.

\begin{table*}[htbp]
	\caption{PERFORMANCE COMPARISON OF OUR METHOD AND SOTA METHODS OVER FOUR HSI DATASETS}\label{H5}
	
	\centering
	\scalebox{0.75}{
		\begin{tabular}{|ccccccccccccc|}			
			\hline
			\multicolumn{1}{|c|}{\multirow{2}{*}{\textbf{Model}}} & \multicolumn{3}{c|}{\textbf{IP}}                                                                & \multicolumn{3}{c|}{\textbf{PU}}                                                         & \multicolumn{3}{c|}{\textbf{SA}} &
			\multicolumn{3}{c|}{\textbf{WHU-Hi-HanChuan}}                                                \\ \cline{2-13} 
			\multicolumn{1}{|c|}{}                       & \multicolumn{1}{c|}{ALL ACC} & \multicolumn{1}{c|}{Known ACC} & \multicolumn{1}{c|}{Unknown ACC} & \multicolumn{1}{c|}{ALL ACC} & \multicolumn{1}{c|}{Known ACC} & \multicolumn{1}{c|}{Unknown ACC} & \multicolumn{1}{c|}{ALL ACC} & \multicolumn{1}{c|}{Known ACC} & \multicolumn{1}{c|}{Unknown ACC} & \multicolumn{1}{c|}{ALL ACC} & \multicolumn{1}{c|}{Known ACC} & \multicolumn{1}{c|}{Unknown ACC}
			 \\ \hline
			\multicolumn{13}{|c|}{1-SHOT Evaluation}                                                                                                                                                                                                                                                                                         \\ \hline

			\multicolumn{1}{|c|}{DFSL}                  & \multicolumn{1}{c|}{53.82}        & \multicolumn{1}{c|}{60.45}          & \multicolumn{1}{c|}{39.55}            & \multicolumn{1}{c|}{53.19}        & \multicolumn{1}{c|}{61.06}          & \multicolumn{1}{c|}{44.61}            & \multicolumn{1}{c|}{50.83}        & \multicolumn{1}{c|}{75.01}          &\multicolumn{1}{c|}{48.71}         & \multicolumn{1}{c|}{42.21} & \multicolumn{1}{c|}{56.55} & \multicolumn{1}{c|}{50.46}         \\ \hline
			\multicolumn{1}{|c|}{MDL4OW}                 & \multicolumn{1}{c|}{42.45}        & \multicolumn{1}{c|}{45.35}          & \multicolumn{1}{c|}{41.59}            & \multicolumn{1}{c|}{51.69}        & \multicolumn{1}{c|}{58.82}          & \multicolumn{1}{c|}{49.82}            & \multicolumn{1}{c|}{56.49}        & \multicolumn{1}{c|}{60.15}          & \multicolumn{1}{c|}{59.53}    & \multicolumn{1}{c|}{43.51} & \multicolumn{1}{c|}{49.26} & \multicolumn{1}{c|}{54.81}      \\ \hline
			\multicolumn{1}{|c|}{OCN}                    & \multicolumn{1}{c|}{55.32}        & \multicolumn{1}{c|}{79.59}          & \multicolumn{1}{c|}{44.98}            & \multicolumn{1}{c|}{53.89}        & \multicolumn{1}{c|}{60.25}          & \multicolumn{1}{c|}{55.08}            & \multicolumn{1}{c|}{62.29}        & \multicolumn{1}{c|}{84.26}          &\multicolumn{1}{c|}{59.55}    & \multicolumn{1}{c|}{47.12} & \multicolumn{1}{c|}{64.53} & \multicolumn{1}{c|}{62.62}          \\ \hline
			\multicolumn{1}{|c|}{MORGAN}                 & \multicolumn{1}{c|}{58.21}        & \multicolumn{1}{c|}{90.35}          & \multicolumn{1}{c|}{44.45}            & \multicolumn{1}{c|}{68.16}        & \multicolumn{1}{c|}{80.59}          & \multicolumn{1}{c|}{52.62}            & \multicolumn{1}{c|}{69.12}        & \multicolumn{1}{c|}{82.16}          & \multicolumn{1}{c|}{55.20}  & \multicolumn{1}{c|}{56.37} & \multicolumn{1}{c|}{67.83} & \multicolumn{1}{c|}{63.64}             \\ \hline
			\multicolumn{1}{|c|}{SSMLP-RPL}                 & \multicolumn{1}{c|}{58.67}        & \multicolumn{1}{c|}{80.18}          & \multicolumn{1}{c|}{40.45}            & \multicolumn{1}{c|}{69.62}        & \multicolumn{1}{c|}{89.30}          & \multicolumn{1}{c|}{56.25}            & \multicolumn{1}{c|}{63.65}        & \multicolumn{1}{c|}{86.52}          &\multicolumn{1}{c|}{56.53}        & \multicolumn{1}{c|}{61.58} & \multicolumn{1}{c|}{66.50} & \multicolumn{1}{c|}{67.90}      \\ \hline
			\multicolumn{1}{|c|}{EVML}                   & \multicolumn{1}{c|}{63.15}        & \multicolumn{1}{c|}{\textbf{91.67}}          & \multicolumn{1}{c|}{52.53}            & \multicolumn{1}{c|}{69.24}        & \multicolumn{1}{c|}{72.83}          & \multicolumn{1}{c|}{61.36}            & \multicolumn{1}{c|}{76.25}        & \multicolumn{1}{c|}{86.28}          &\multicolumn{1}{c|}{61.64}     & \multicolumn{1}{c|}{64.27} & \multicolumn{1}{c|}{69.69} & \multicolumn{1}{c|}{62.19
			}         \\ \hline
			\multicolumn{1}{|c|}{ours}                   & \multicolumn{1}{c|}{	\textbf{66.66}}        & \multicolumn{1}{c|}{72.28}          & \multicolumn{1}{c|}{\textbf{54.94}}            & \multicolumn{1}{c|}{\textbf{74.34}}        & \multicolumn{1}{c|}{\textbf{93.59}}          & \multicolumn{1}{c|}{\textbf{62.71}}            & \multicolumn{1}{c|}{\textbf{78.58}}        & \multicolumn{1}{c|}{\textbf{95.51}}          & \multicolumn{1}{c|}{\textbf{66.45}}      & \multicolumn{1}{c|}{\textbf{65.69}} & \multicolumn{1}{c|}{\textbf{70.93}} & \multicolumn{1}{c|}{\textbf{69.12}}         \\ \hline
			\multicolumn{13}{|c|}{5-SHOT Evaluation}                                                                                                                                                                                                                                                                                         \\ \hline

			\multicolumn{1}{|c|}{DFSL}                  & \multicolumn{1}{c|}{55.45}        & \multicolumn{1}{c|}{73.05}          & \multicolumn{1}{c|}{41.68}            & \multicolumn{1}{c|}{50.56}        & \multicolumn{1}{c|}{56.27}          & \multicolumn{1}{c|}{39.83}            & \multicolumn{1}{c|}{64.57}        & \multicolumn{1}{c|}{81.22}          &  \multicolumn{1}{c|}{56.35}     & \multicolumn{1}{c|}{50.13} & \multicolumn{1}{c|}{59.55} & \multicolumn{1}{c|}{54.56}         \\ \hline
			\multicolumn{1}{|c|}{MDL4OW}                 & \multicolumn{1}{c|}{50.39}        & \multicolumn{1}{c|}{52.83}          & \multicolumn{1}{c|}{46.62}            & \multicolumn{1}{c|}{59.52}        & \multicolumn{1}{c|}{65.24}          & \multicolumn{1}{c|}{53.7}            & \multicolumn{1}{c|}{66.16}        & \multicolumn{1}{c|}{76.29}          &    \multicolumn{1}{c|}{56.7}     & \multicolumn{1}{c|}{52.83} & \multicolumn{1}{c|}{55.83} & \multicolumn{1}{c|}{56.53}      \\ \hline
			\multicolumn{1}{|c|}{OCN}                    & \multicolumn{1}{c|}{62.14}        & \multicolumn{1}{c|}{93.26}          & \multicolumn{1}{c|}{49.51}            & \multicolumn{1}{c|}{62.99}        & \multicolumn{1}{c|}{73.44}          & \multicolumn{1}{c|}{56.07}            & \multicolumn{1}{c|}{79.36}        & \multicolumn{1}{c|}{86.83}          &\multicolumn{1}{c|}{67.29}    & \multicolumn{1}{c|}{52.43} & \multicolumn{1}{c|}{69.30} & \multicolumn{1}{c|}{74.51}           \\ \hline
			\multicolumn{1}{|c|}{MORGAN}                 & \multicolumn{1}{c|}{60.07}        & \multicolumn{1}{c|}{92.48}          & \multicolumn{1}{c|}{42.53}            & \multicolumn{1}{c|}{70.40}        & \multicolumn{1}{c|}{83.86}          & \multicolumn{1}{c|}{56.67}            & \multicolumn{1}{c|}{76.11}        & \multicolumn{1}{c|}{88.47}          & \multicolumn{1}{c|}{55.18}     & \multicolumn{1}{c|}{63.26} & \multicolumn{1}{c|}{71.31} & \multicolumn{1}{c|}{70.39}         \\ \hline
			\multicolumn{1}{|c|}{SSMLP-RPL}                 & \multicolumn{1}{c|}{72.8}        & \multicolumn{1}{c|}{88.36}          & \multicolumn{1}{c|}{63.21}            & \multicolumn{1}{c|}{72.5}        & \multicolumn{1}{c|}{97.64}          & \multicolumn{1}{c|}{57.6}            & \multicolumn{1}{c|}{72.67}        & \multicolumn{1}{c|}{96.59}          & \multicolumn{1}{c|}{68.91}      & \multicolumn{1}{c|}{66.01}  & \multicolumn{1}{c|}{85.46} & \multicolumn{1}{c|}{57.91}        \\ \hline
			\multicolumn{1}{|c|}{EVML}                   & \multicolumn{1}{c|}{70.12}        & \multicolumn{1}{c|}{\textbf{93.33}}          & \multicolumn{1}{c|}{54.62}            & \multicolumn{1}{c|}{81.67}        & \multicolumn{1}{c|}{88.36}          & \multicolumn{1}{c|}{65.01}            & \multicolumn{1}{c|}{81.50}        & \multicolumn{1}{c|}{95.06}          &\multicolumn{1}{c|}{72.09}   & \multicolumn{1}{c|}{70.62} & \multicolumn{1}{c|}{71.07} & \multicolumn{1}{c|}{77.90}           \\ \hline
			\multicolumn{1}{|c|}{ours}                   & \multicolumn{1}{c|}{\textbf{78.14}}        & \multicolumn{1}{c|}{87.68}          & \multicolumn{1}{c|}{\textbf{70.39}}            & \multicolumn{1}{c|}{\textbf{84.27}}        & \multicolumn{1}{c|}{\textbf{98.08}}          & \multicolumn{1}{c|}{\textbf{68.2}}            & \multicolumn{1}{c|}{\textbf{83.79}}        & \multicolumn{1}{c|}{\textbf{97.02}}          & \multicolumn{1}{c|}{\textbf{73.24}}    & \multicolumn{1}{c|}{\textbf{89.41}} & \multicolumn{1}{c|}{\textbf{88.98}} & \multicolumn{1}{c|}{\textbf{82.37}}         \\ \hline
	\end{tabular}}
\end{table*}

Table \ref{H5} provides a comparative analysis of our model's performance against other models under 1-shot and 5-shot settings. The comparison results indicate that among these existing methods, SSMLP-RPL, OCN, MORGAN and EVML perform known class classification well for both 5-shot and 1-shot scenarios in terms of known ACC metrics. And in terms of the Unknown ACC, EVML exhibits marginally superior performance.  But as shown in Table \ref{H5}, our model outperforms all these models in most cases. For example, compared with the optimal result of these existing methods under the 5-shot setup, our approach surpasses other methods by 1.9\% in terms of known ACC on SA dataset, exceeds by 11\% in terms of known ACC on WHU-Hi-HanChuan dataset, improves by 3\% in terms of ALL ACC on PU dataset, and enhances by 16\% in terms of unknown ACC on IP dataset. At the same time, under the 1-shot setup, our model achieves a almost 5\% higher unknown ACC on SA dataset, a 13\% increase in terms of known ACC on PU dataset, and a 3\% enhancement in terms of ALL ACC on IP dataset. These results demonstrate that while maintaining high accuracy in known class classification, our model posses the ability to recognize potential unknown classes in HSIs. 

\begin{figure*}[htbp]
	\centering
	\subfigure[]{
		\includegraphics[width=0.16\textwidth]{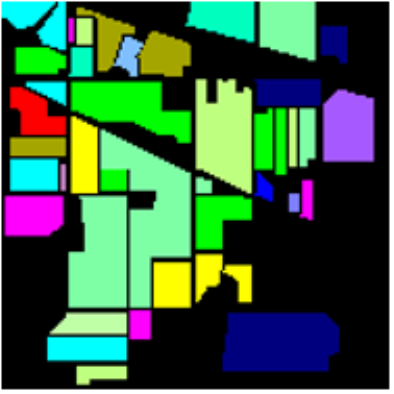}
	}
	\quad
	\subfigure[]{
		\includegraphics[width=0.16\textwidth]{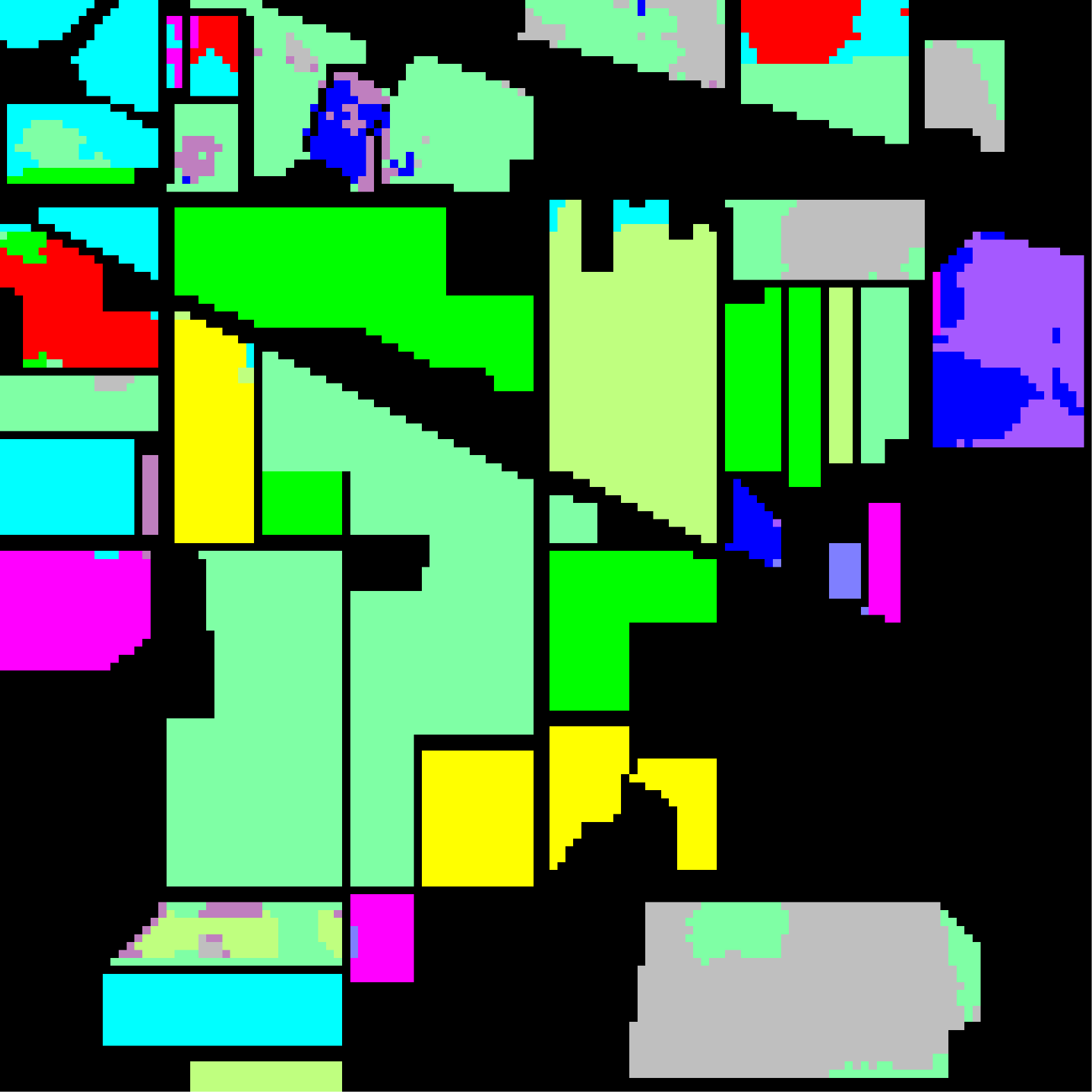}
	}
	\quad
	\subfigure[]{
		\includegraphics[width=0.16\textwidth]{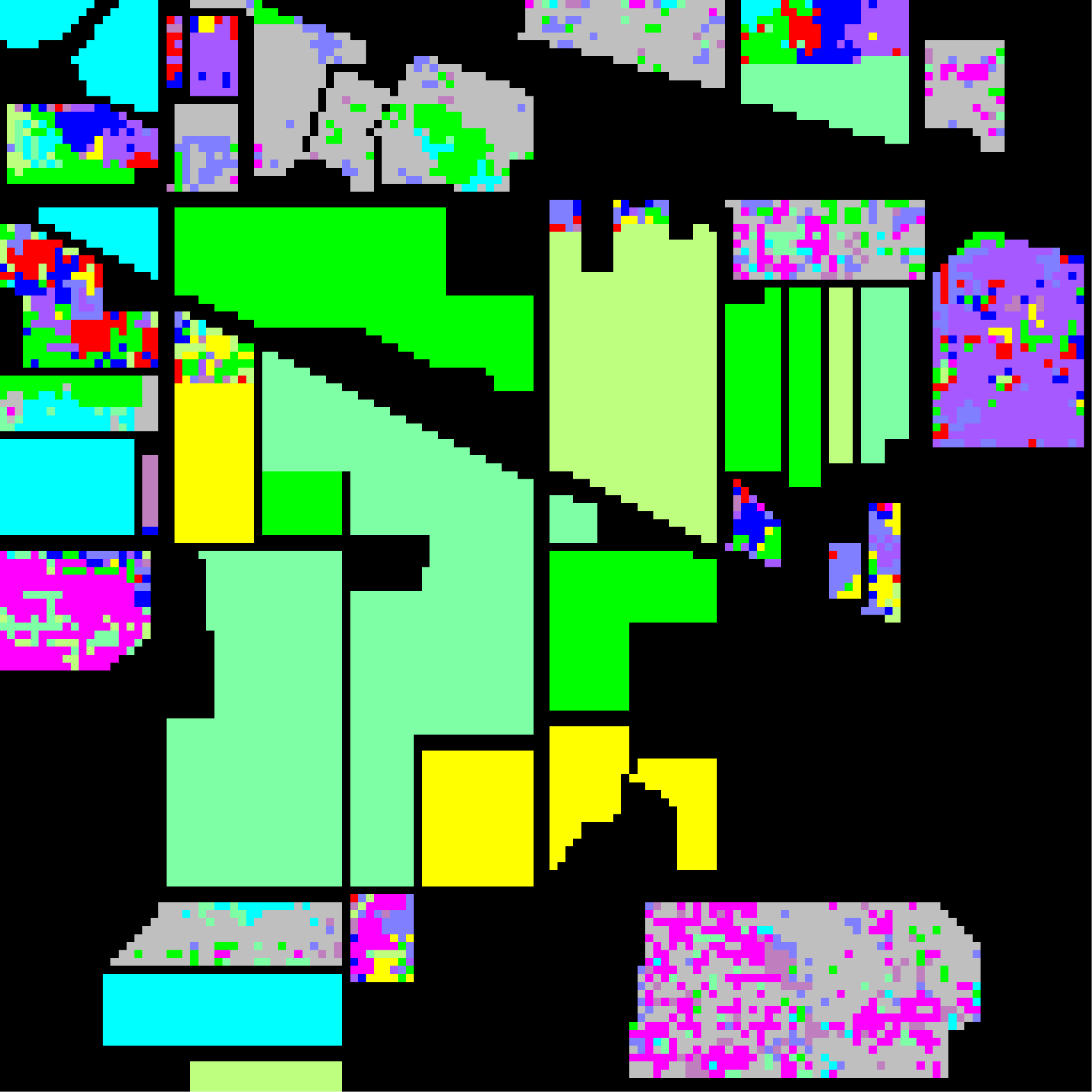}
	}
	\quad
   \subfigure[]{
	   \includegraphics[width=0.16\textwidth]{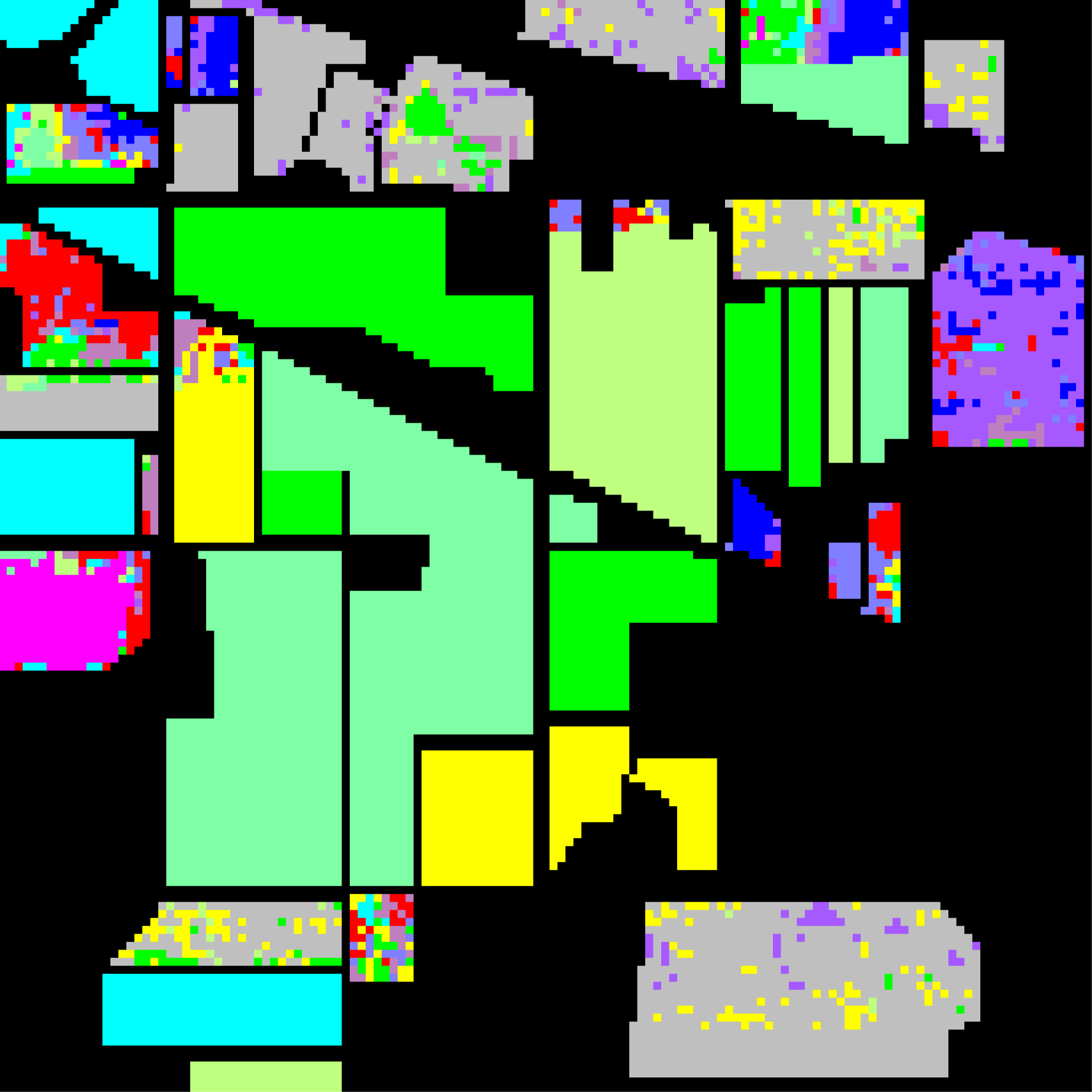}
    }
	\quad
    \subfigure[]{
     	\includegraphics[width=0.16\textwidth]{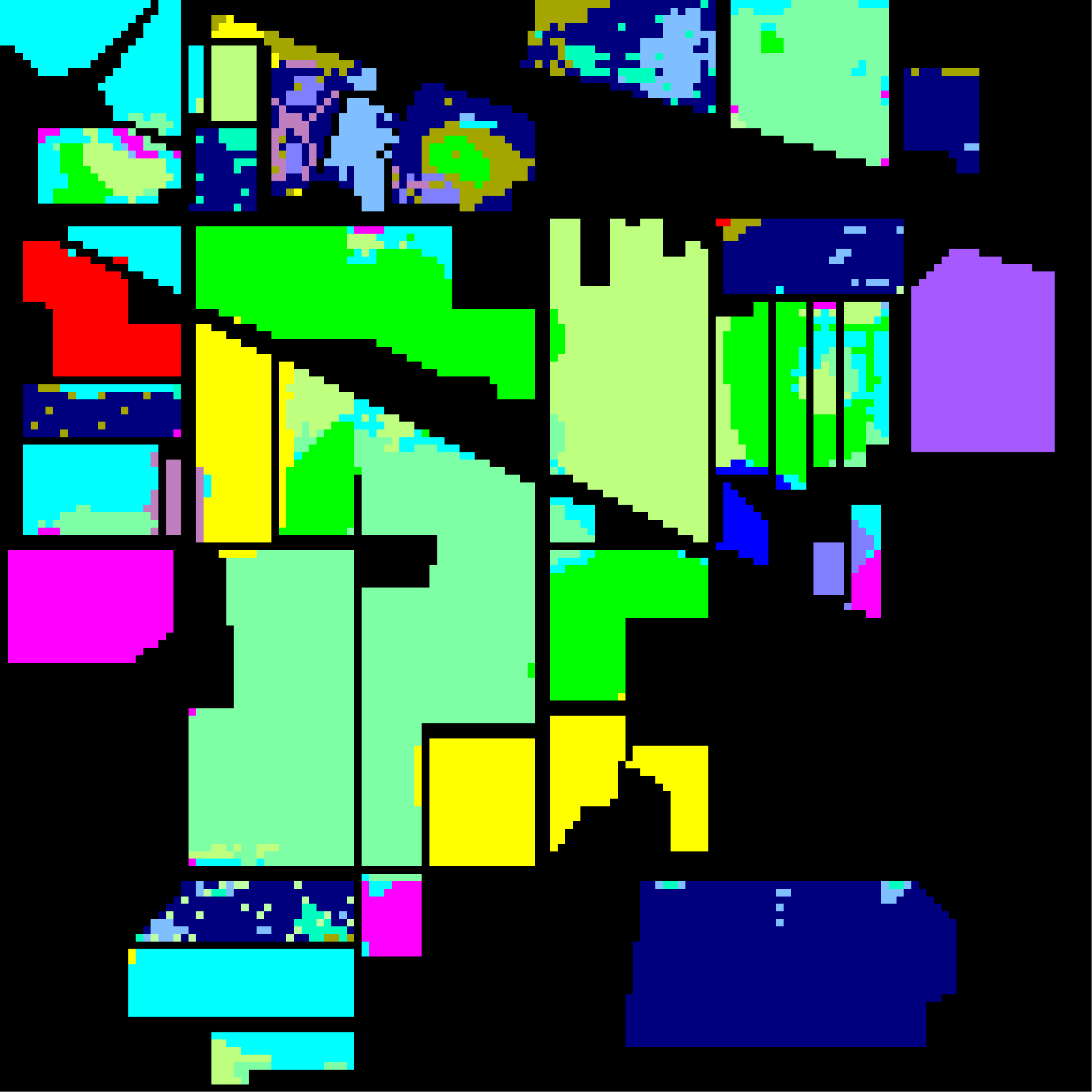}
    }
	\caption{ Data visualization and classification maps using different methods, including: (a) ground-truth map of IP dataset, (b) EVML, (c) MORGAN, (d) OCN, (e) ours}
	\label{tu1}
\end{figure*}
\begin{figure*}[htbp]

	\centering
	\subfigure[]{
		\includegraphics[width=0.165\textwidth]{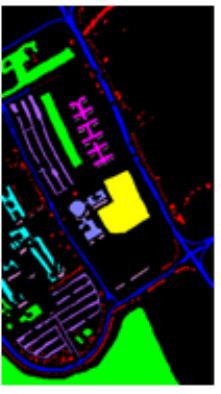}
	}
	\quad
	\subfigure[]{
		\includegraphics[width=0.16\textwidth]{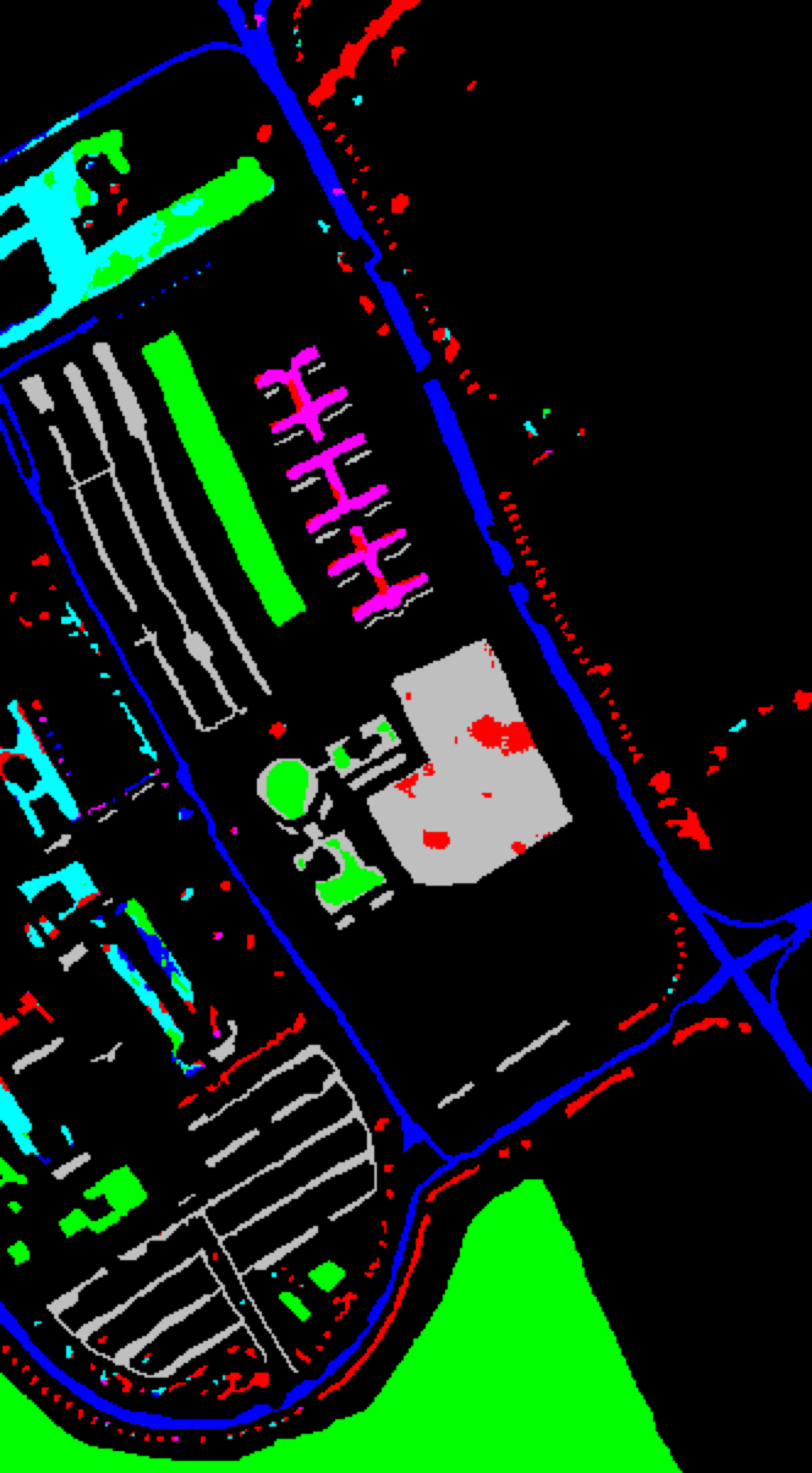}
	}
	\quad
	\subfigure[]{
		\includegraphics[width=0.16\textwidth]{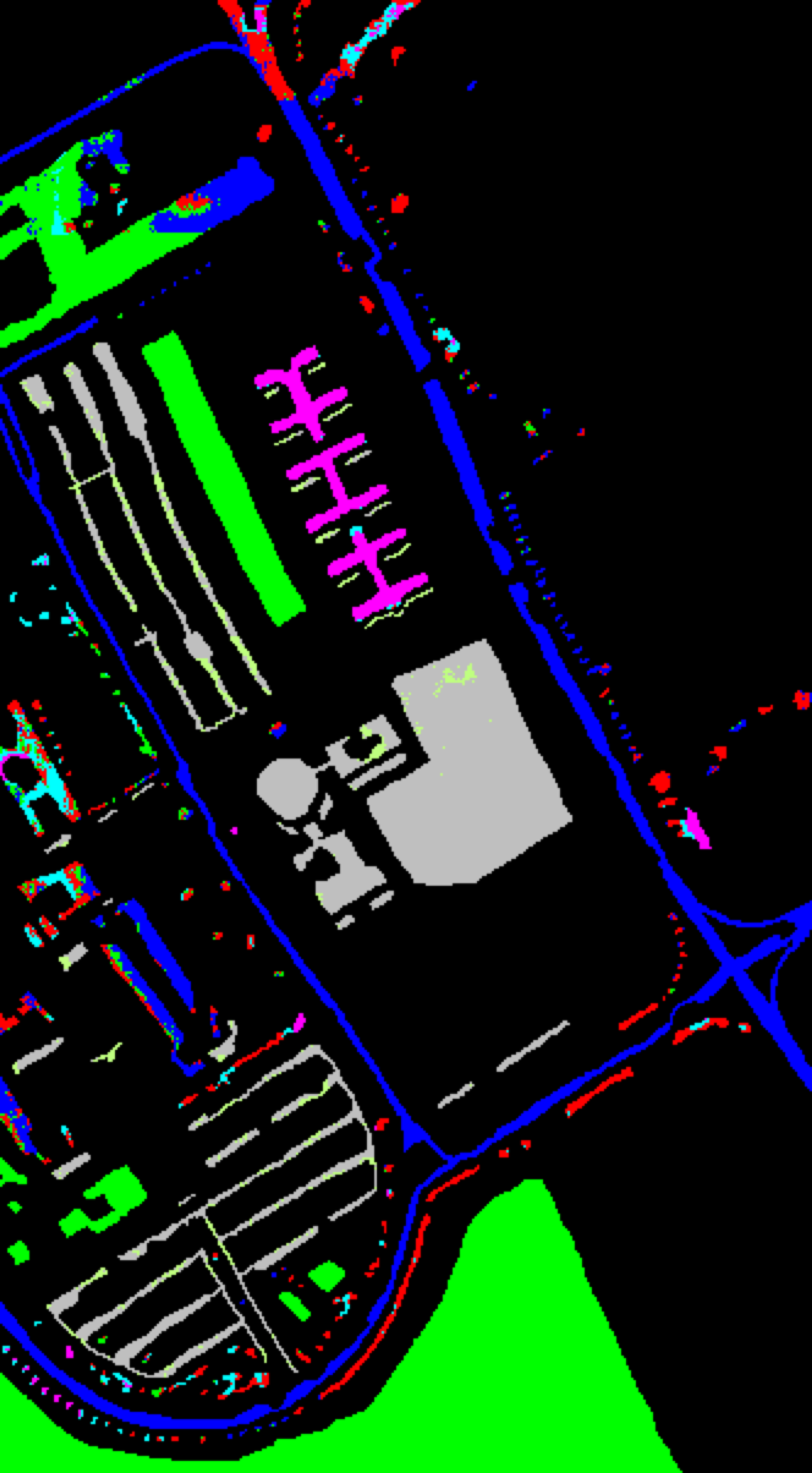}
	}
	\quad
   \subfigure[]{
	   \includegraphics[width=0.16\textwidth]{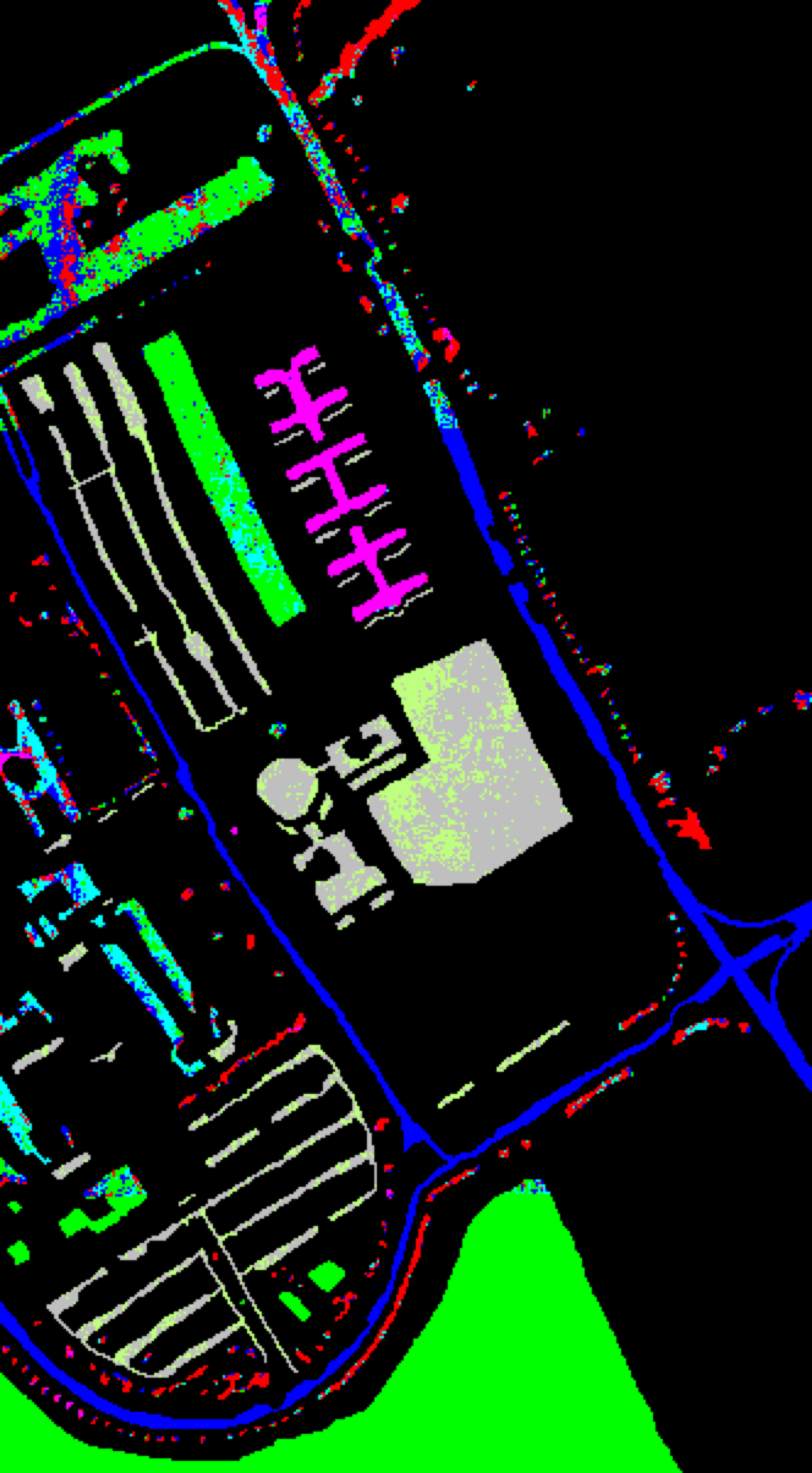}
    }
	\quad
    \subfigure[]{
     	\includegraphics[width=0.16\textwidth]{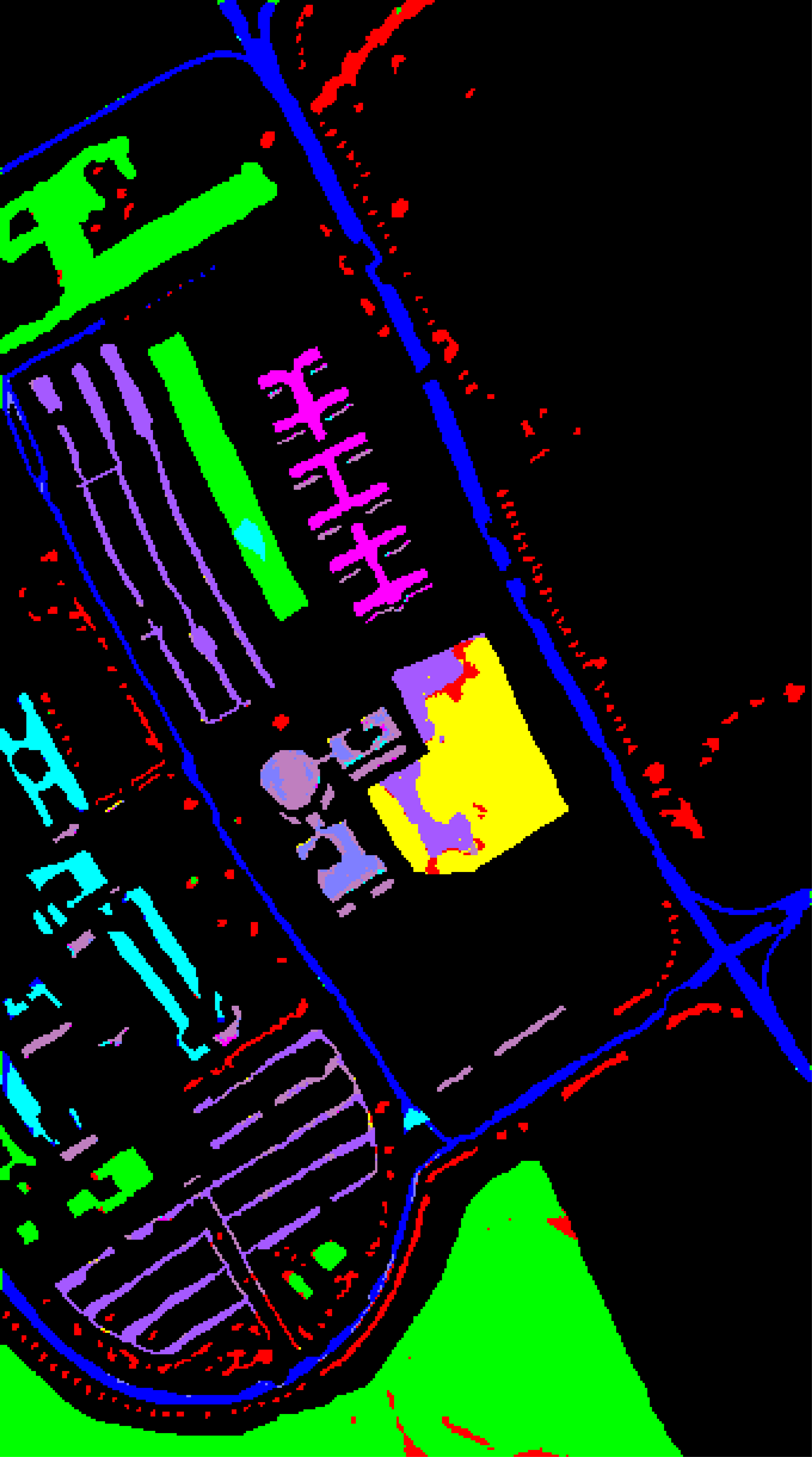}
    }
	\caption{ Data visualization and classification maps using different methods, including: (a) ground-turth map of PU dataset, (b) EVML, (c) MORGAN, (d) OCN, (e) ours}
	\label{tu2}
 \end{figure*}
\begin{figure*}[htbp]
	\centering
	\subfigure[]{
		\includegraphics[width=0.16\textwidth]{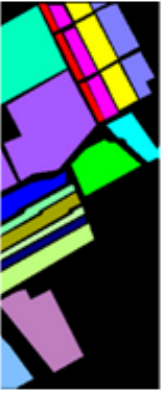}
	}
	\quad
	\subfigure[]{
		\includegraphics[width=0.16\textwidth]{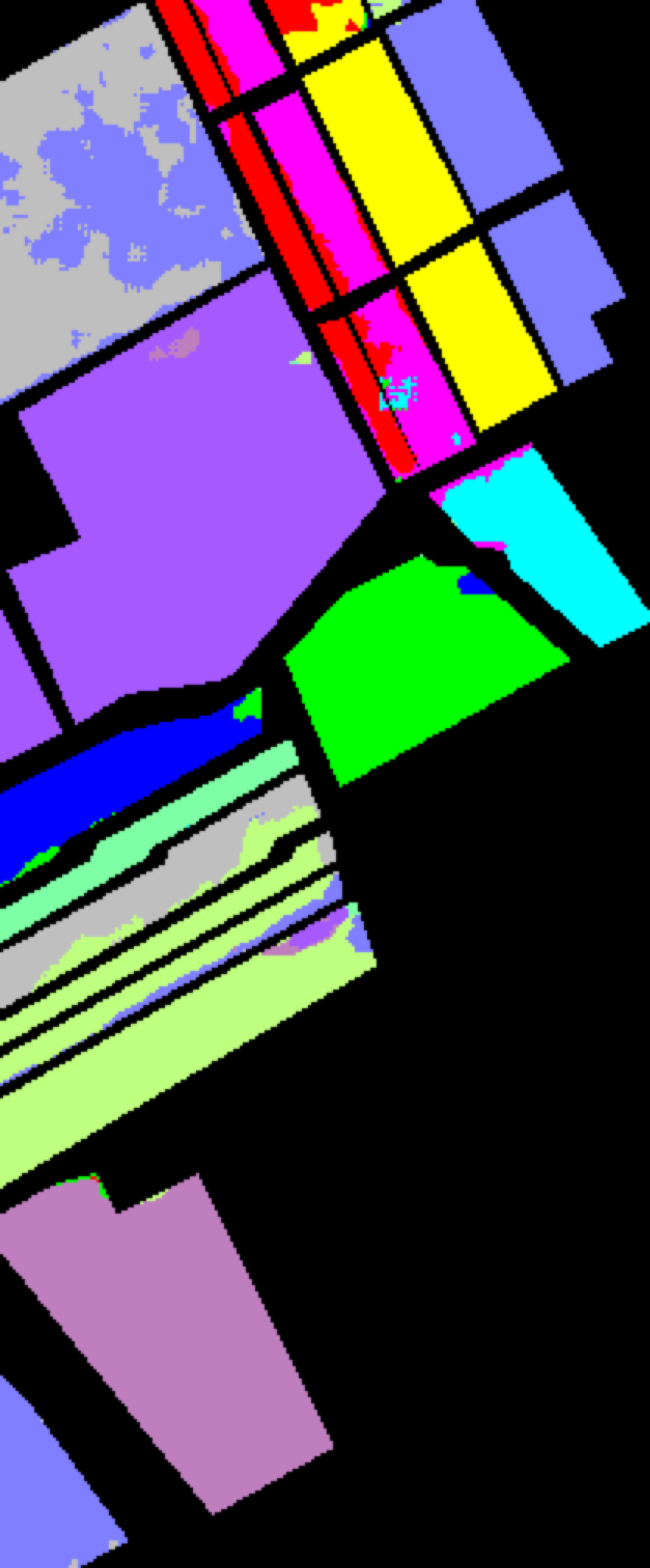}
	}
	\quad
	\subfigure[]{
		\includegraphics[width=0.16\textwidth]{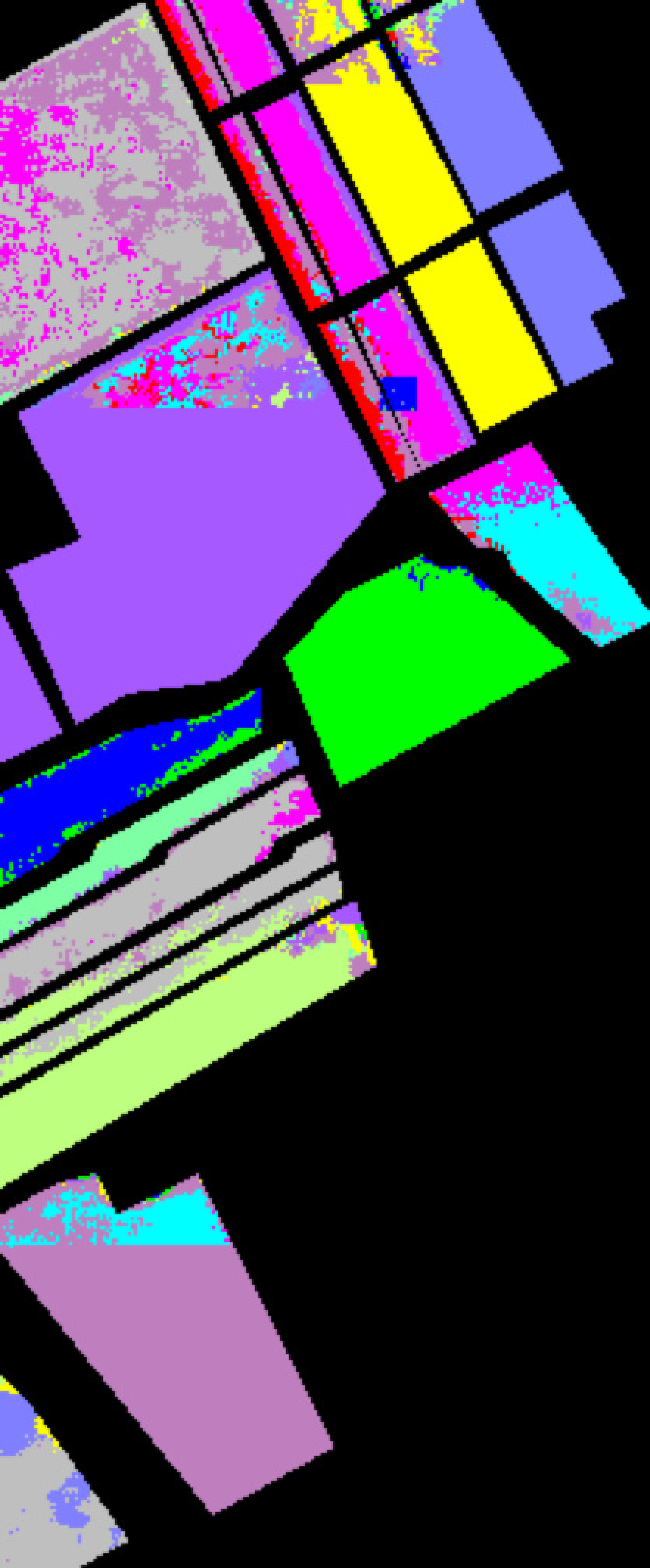}
	}
	\quad
   \subfigure[]{
	   \includegraphics[width=0.16\textwidth]{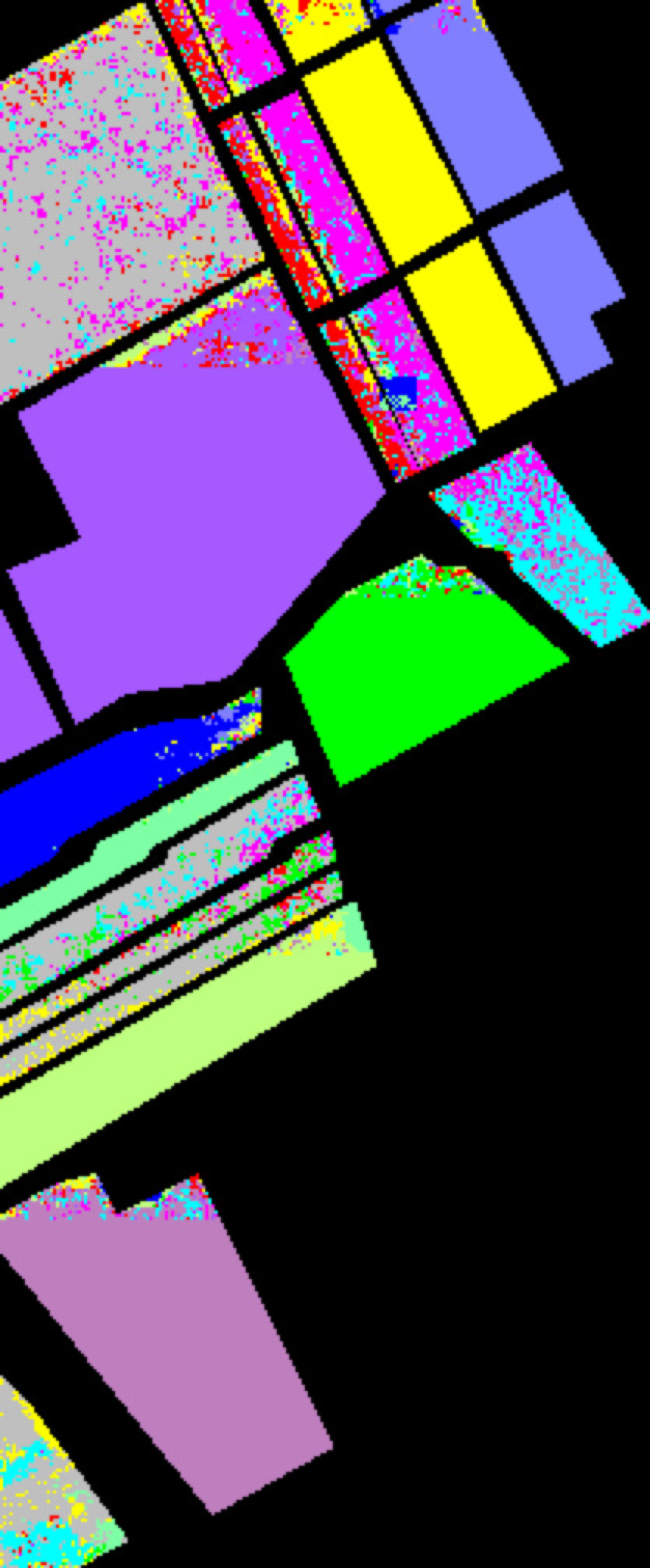}
    }
	\quad
    \subfigure[]{
     	\includegraphics[width=0.165\textwidth]{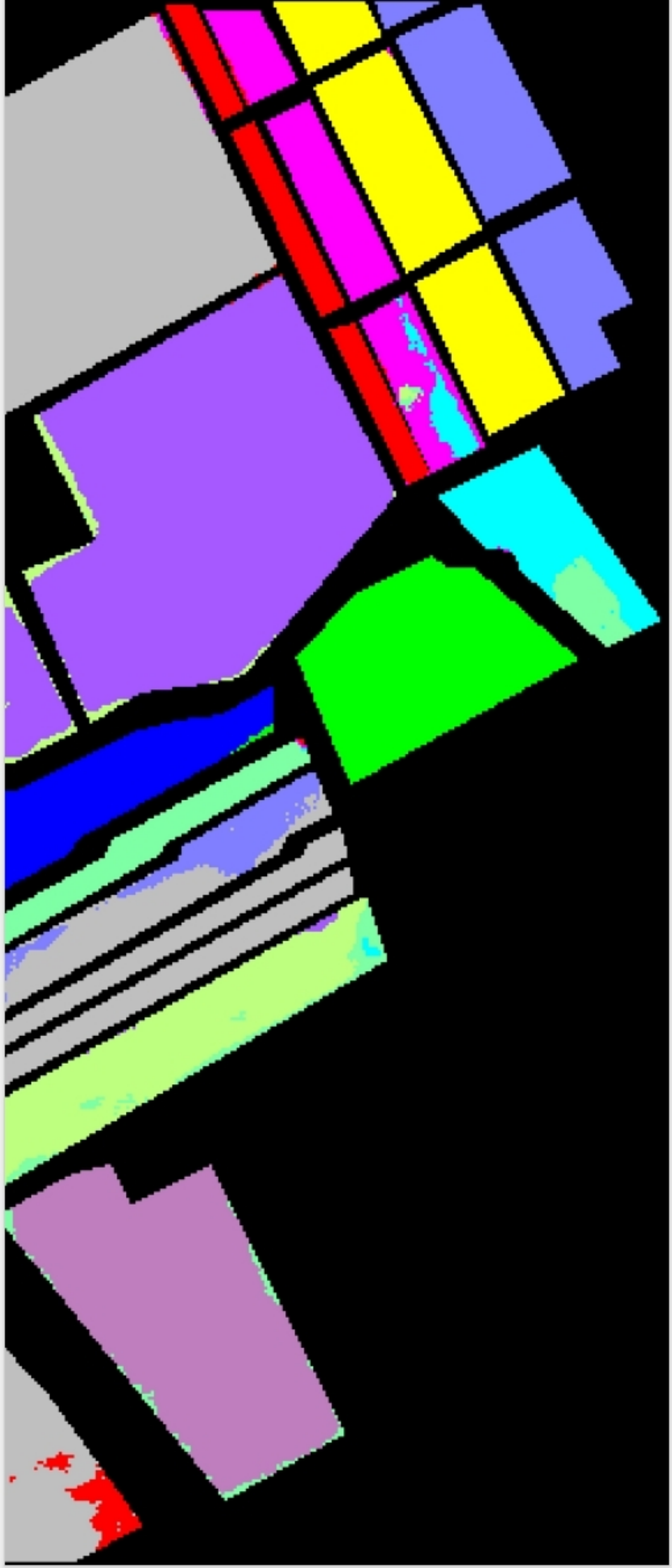}
    }

	\caption{Data visualization and classification maps using different methods, including: (a) ground reality map of SA, (b) EVML, (c) MORGAN, (d) OCN, (e) ours}
	\label{tu3}
\end{figure*}
\begin{figure*}[htbp]
	\centering
	\subfigure[]{
		\includegraphics[width=0.1515\textwidth]{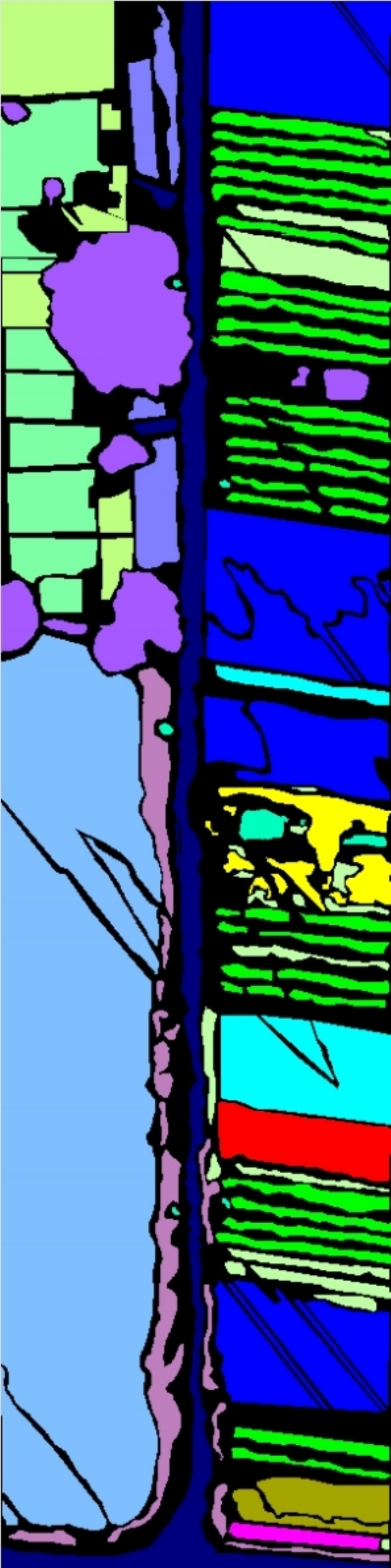}
	}
	\quad
	\subfigure[]{
		\includegraphics[width=0.148\textwidth]{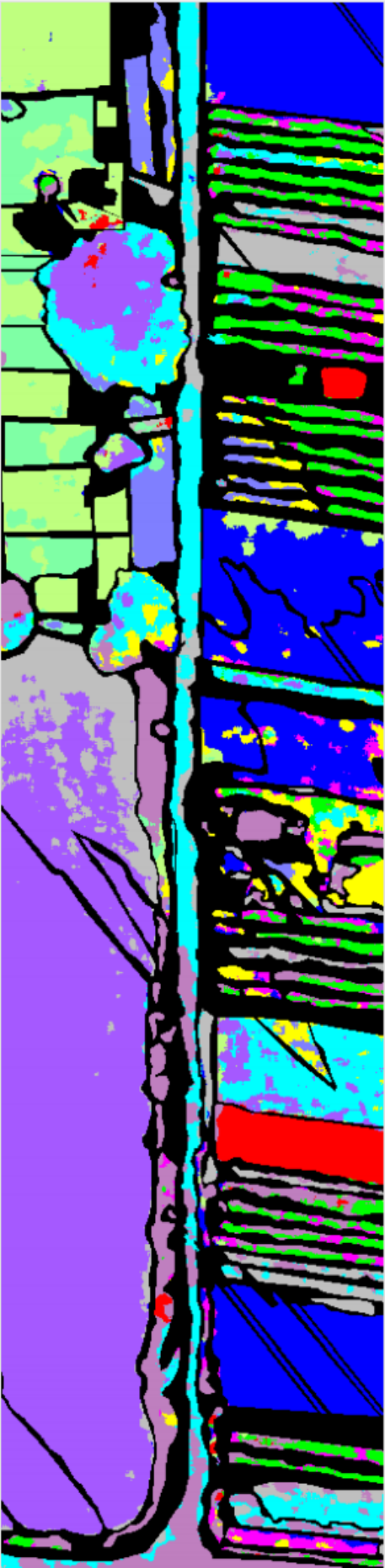}
	}
	\quad
	\subfigure[]{
		\includegraphics[width=0.149\textwidth]{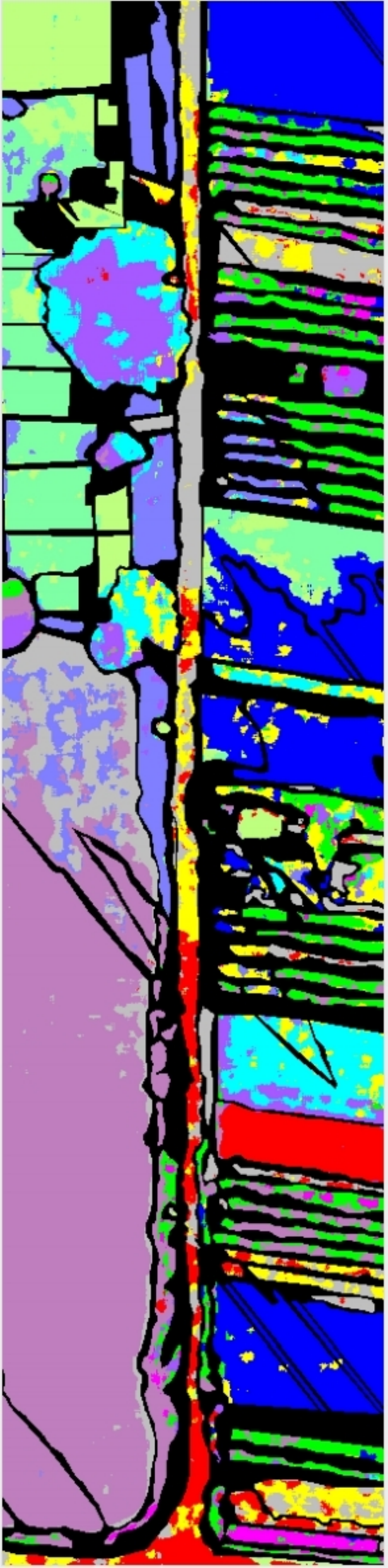}
	} 
	\quad
	\subfigure[]{
		\includegraphics[width=0.146\textwidth]{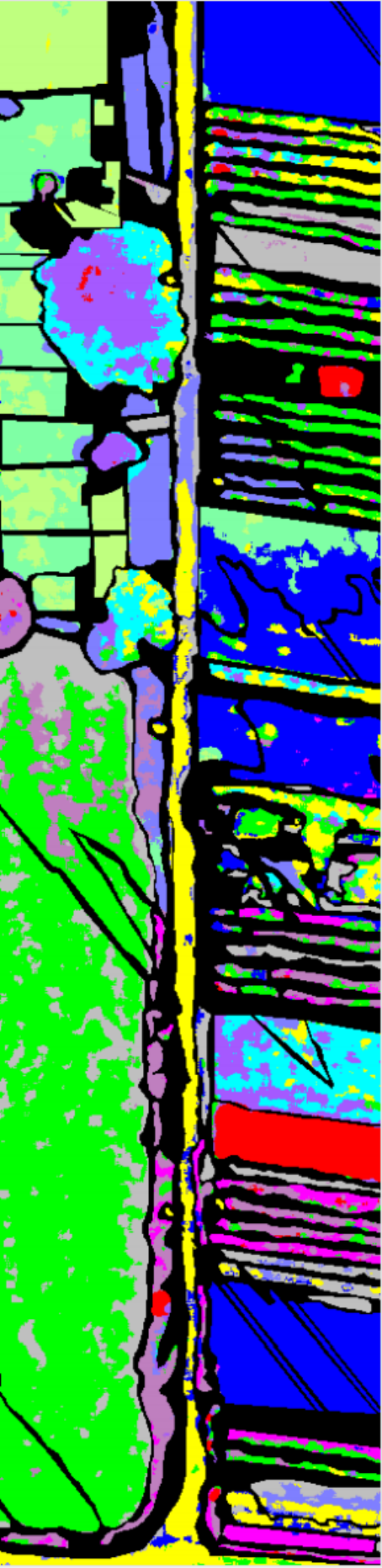}
	}
	\quad
	\subfigure[]{
		\includegraphics[width=0.15\textwidth]{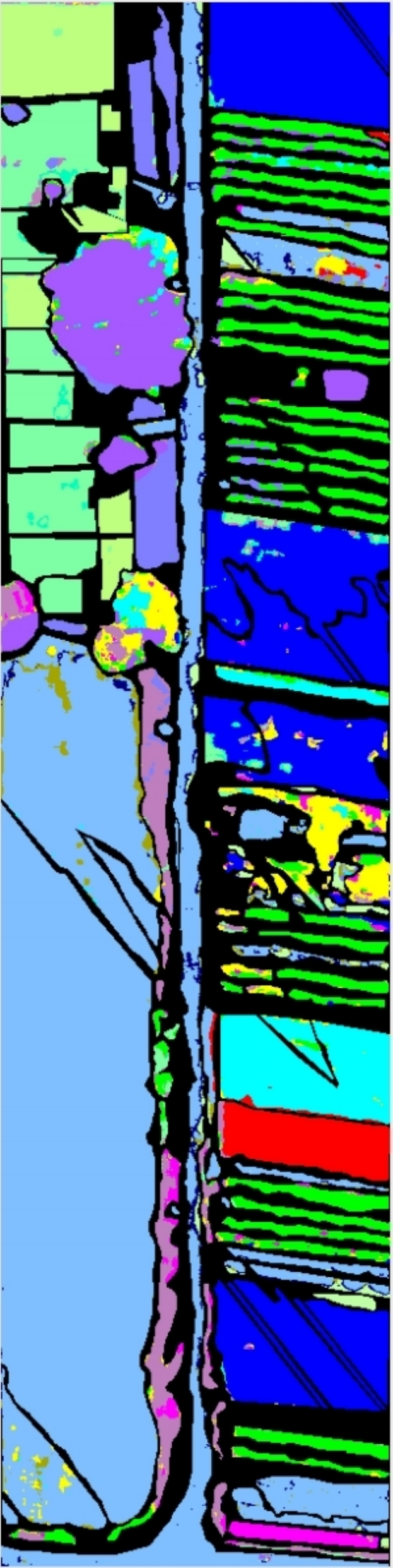}
	}
	
	\caption{Data visualization and classification maps using different methods, including: (a) ground reality map of WHU-Hi-HanChuan, (b) EVML, (c) MORGAN, (d) OCN, (e) ours}
	\label{tu4}
\end{figure*}

We selected the classification maps from EVML, MORAN, OCN, and our method to further examine comparative results. These classification maps can be seen in Fig. \ref{tu1}, \ref{tu2}, \ref{tu3} and \ref{tu4}. For EVML, MORAN and OCN which reject the unknow classes while classifying known classes, we use the gray color to indicate the unknown classes in the classification maps. From these classification maps, it can be observed that our model can not only reject unknown classes but also further  finely cluster them (i.e., the last 5 classes in IP dataset, the last 5 classes in SA dataset, the last 4 classes in PU dataset, and the last 5 classses in WHU-Hi-HanChuan dataset).

Although the model shows potential in detecting unknown categories and effectively aligning them with the actual category hierarchy, its application on the IP dataset reveals limitations: the match between the prototype group labels and the true unknown category labels is unsatisfactory. In contrast, the model demonstrates clear advantages on the SA and PU datasets.

\subsection{Visualization of the Feature Space}
\begin{figure*}[htbp]
	\centering
	\subfigure[]{
		\includegraphics[width=0.6\textwidth]{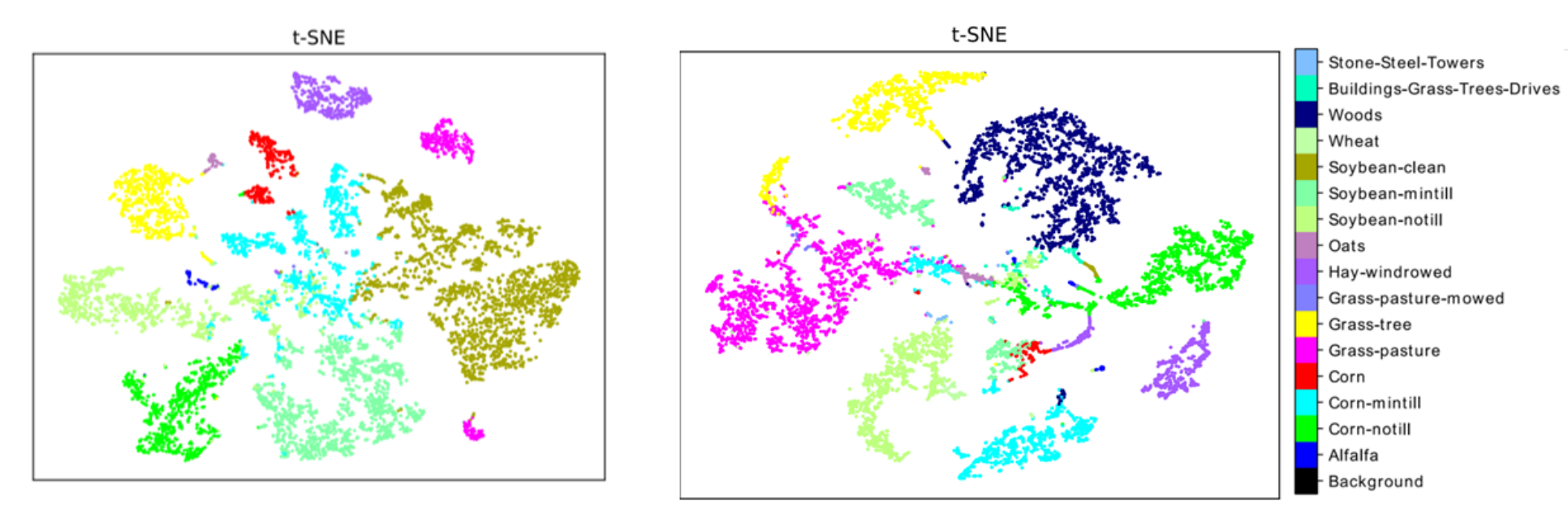}
	}
	\quad
	\subfigure[]{
		\includegraphics[width=0.5\textwidth]{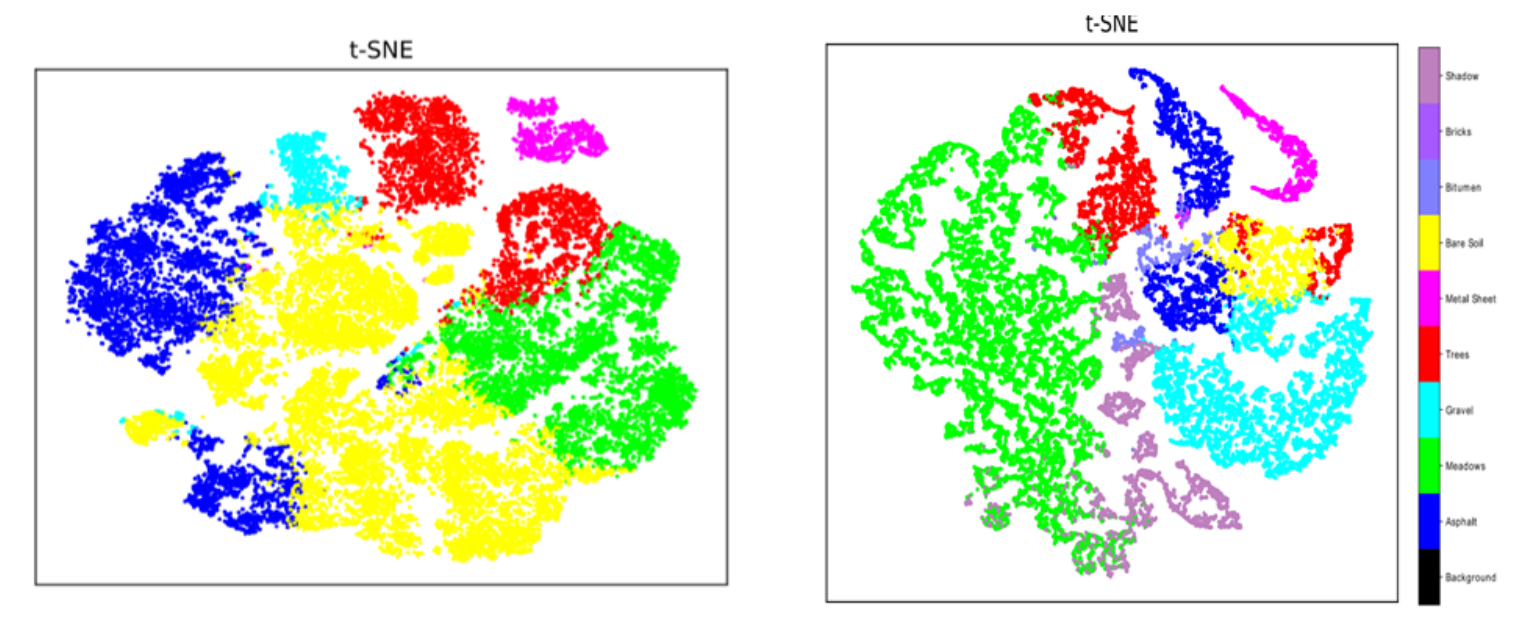}
	}
	\quad
	\subfigure[]{
		\includegraphics[width=0.7\textwidth]{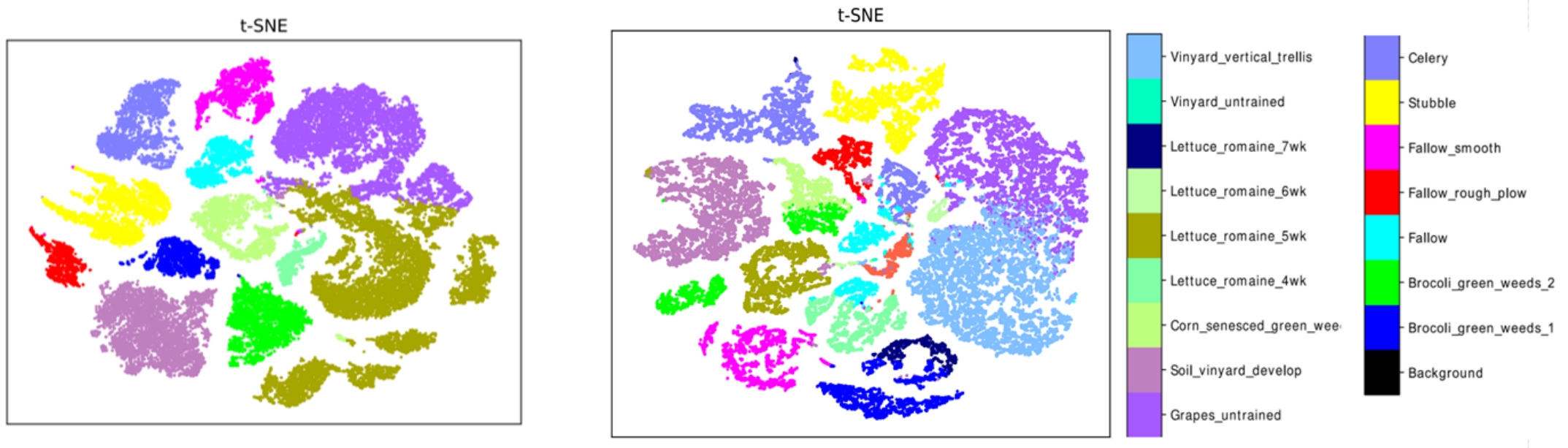}
	}

	\caption{illustrates the learned few-sample, bi-level contrastive learning feature representations for three benchmark HSI datasets: (a)-right: Indian Pines, (b)-right: University of Pavia, and (c)-right: Salinas. It also shows the t-SNE visualizations of land cover categories centered around class anchors for (a)-left Indian Pines, (b)-left University of Pavia, and (c)-left Salinas, with colors representing different categories.}\label{H14}
\end{figure*}

In Fig. \ref{H14}, we employ the t-SNE technique to visualize the embedding features extracted by the feature extractor. This visualization reveals the structural features of the latent space. In the figure, the 12th class (bright silver) of IP data set and SA data set is unknown, while the 6th class (yellow) of PU data set is unknown. As shown in Figures \ref{H14}-left, under the guidance of class anchors, the embedding features of the known classes form clear clusters around the class anchors in the logit space, effectively isolating the samples of unknown classes. Building on this, as depicted in Fig. \ref{H14}-right, we further explore the unknown classes by grouping prototypes based on pairwise similarities between prototypes and their groups. This process aims to form compact, distinct clusters and align these clusters with the actual class hierarchy for efficient recognition of unknown classes.

\subsection{Ablation Experiments}

There is a set of loss items in the objective function of the proposed method, including open set classification loss $L_{osc}$, class anchor loss $L_{ca}$, prototype similarity loss $L_{ps}$, prototype group similarity loss $L_{pgs}$, known class discovery loss $L_{kcd}$, and prototype regularization loss $L_{reg}$. To elucidate the importance of these components, ablation studies were conducted by sequentially omitting each term from the objective function. Table \ref{H45} provides an in-depth analysis of the contributions of these loss function components on IP dataset.

\begin{table}[H]
	\centering
	\caption{Ablation analysis of the objective function components on IP dataset}\label{H45}
	\begin{tabular}{lccc}
		\hline
		\textbf{Methods} & \textbf{Closed OA} & \textbf{Open OA} & \textbf{AUROC} \\ \hline
		w/o $L_{pair}$ & 83.16 & 80.16 & 79.63 \\
		w/o $L_{ce}^{sim}$ & 87.49 & 82.81 & 81.52 \\
		w/o\ $L_{ce}^{pseudo}$ & 92.29 & 90.48 & 90.66 \\
		w/o $L_{reg}$ & 95.18 & 92.19 & 95 \\
		ours & 97.78 & 96.05 & 98.06 \\ \hline
	\end{tabular}
\end{table}
It is evident that each component synergistically contributes to enhancing the performance of our complete model, surpassing the ablated counterparts across all considered metrics.  This emphasizes the indispensable role that each loss function component plays in achieving state-of-the-art model performance. Notably, the ablation results highlight the pivotal role of the open set classification loss $L_{osc}$ and the class anchor loss $L_{ca}$.


\subsection{Benefits of Multiple Prototypes and Class Anchor}

\begin{figure*}[htbp]
	\centering
	\subfigure[]{
		\includegraphics[width=0.3\textwidth]{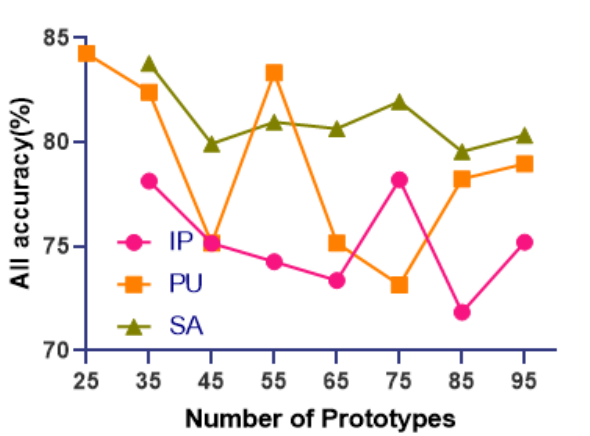}\label{H20}
	}
	\quad
	\subfigure[]{
		\includegraphics[width=0.3\textwidth]{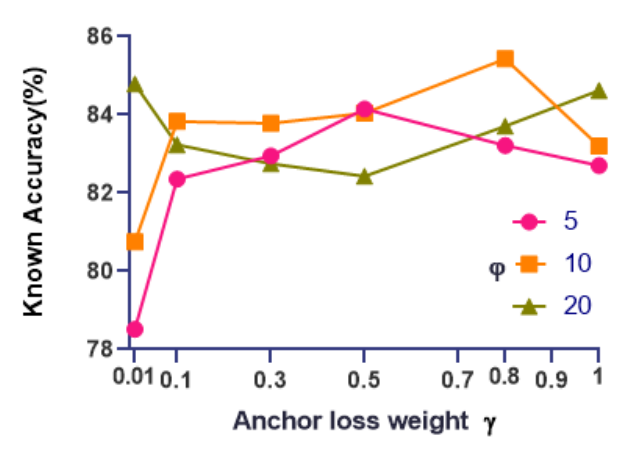}\label{H21}
	}
	\caption{(a): The impact analysis of the number of prototypes on IP dataset. (b) The impact evaluation of of the hyper-parameters of $\varphi$  shown in Eq. \eqref{eq:4} and $\gamma$ shown in Eq. \eqref{eq:ca}}
\end{figure*}

\begin{table*}[htbp]
	\caption{The number of discovered hyperspectral classes under 1-shot and 5-shot across three datasets}\label{H10}
	
	\centering
	\scalebox{1}{
		\begin{tabular}{|cccccccccccc|}			
			\hline
			\multicolumn{1}{|c|}{\multirow{2}{*}{\textbf{Methods}}} & \multicolumn{2}{c|}{\textbf{IP}}                                                                & \multicolumn{2}{c|}{\textbf{PU}}                                                         & \multicolumn{2}{c|}{\textbf{SA}}  &
			     \multicolumn{2}{c|}{\textbf{WHU-Hi-HanChuan}}                                 \\ \cline{2-9} 
			\multicolumn{1}{|c|}{}                       & \multicolumn{1}{c|}{True Num} & \multicolumn{1}{c|}{Pred Num} & \multicolumn{1}{c|}{True Num} & \multicolumn{1}{c|}{Pred Num} & \multicolumn{1}{c|}{True Num} & \multicolumn{1}{c|}{Pred Num}  & \multicolumn{1}{c|}{True Num} & \multicolumn{1}{c|}{Pred Num}
			\\ \hline
			\multicolumn{1}{|c|}{\textbf{1-shot}}                 & \multicolumn{1}{c|}{16}        & \multicolumn{1}{c|}{17}          & \multicolumn{1}{c|}{9}            & \multicolumn{1}{c|}{15}        & \multicolumn{1}{c|}{16}          & \multicolumn{1}{c|}{21}    & \multicolumn{1}{c|}{16}          & \multicolumn{1}{c|}{21}               
			 \\ \hline
			 \multicolumn{1}{|c|}{\textbf{5-shot}}                 & \multicolumn{1}{c|}{16}        & \multicolumn{1}{c|}{17}          & \multicolumn{1}{c|}{9}            & \multicolumn{1}{c|}{10}        & \multicolumn{1}{c|}{16}          & \multicolumn{1}{c|}{17}     & \multicolumn{1}{c|}{16}          & \multicolumn{1}{c|}{18}               
			 \\ \hline

	\end{tabular}}
\end{table*}

Fig. \ref{H20} demonstrates the overall class performance evaluated across various datasets using different numbers of prototypes, ranging from 25 to 95. It is observed that the classification performance does not increase with an excess of prototypes beyond the actual number of classes, potentially leading to diminished effectiveness. Moreover, Table \ref{H10} shows the real number of classes and the number of discovered classes across four datasets. The data in Table \ref{H10} indicate  that compared with 1-shot setting, the number of classes can be better estimated under the 5-shot setup.

As illustrated in Fig. \ref{H21}, we evaluate impact of the hyper-parameters of $\varphi$ and $\gamma$. The former adjusts the scale of the initialized  class anchors shown in Eq. \eqref{eq:4}.  The latter is to control the impact of the penalty item applied on the Euclidean distance between the predicted logits of the input samples and their true class anchors, shown in Eq. \eqref{eq:ca}.  From Fig. \ref{H21}, it is observable that hyper-parameters of $\varphi$ impacts the optimal location and peak accuracy differently. When $\varphi=10$, the optimal performance is achieved where  $\gamma=0.8$. Therefore, we set $\varphi$ to 10 and $\gamma$ to 0.8 in our experiments.



\section{Conclusion}\label{H2}
In open-set setting, the samples to be classified may be not all from the classes that have been seen. This poses challenges for HSI classification, particularly with limited labeled samples. Different from current methods which focus on distinguishing unknown class samples from known class samples and rejecting them, this paper proposes a novel approach to discover the potential unknown classes while classifying the known class samples. It adopts a class anchor based classification strategy to learn to distinguish unknown classes from known classes. Through expanding the original dimensional logit space to include an additional dimension specifically for representing unknown class, the classifier can recognize the unknow class samples while classifying these known class samples. In order to further discover the potential unknown classes among the samples,  a prototype learning and clustering method is used, which initializes multiple trainable prototypes whose number is significantly greater than that of actual unknown classes and then clusters the prototypes into different groups representing the potential unknown classes. Based on three HSI datasets which have been widely used, extensive experiments have been done. And experimental results have shown the superior performance of the proposed method in addressing open-set few-shot HSI classification when compared with related methods.

\bibliographystyle{IEEEtran}

\bibliography{new}

\vfill

\end{document}